\documentclass[10pt,journal,compsoc]{IEEEtran}
\usepackage{amssymb}
\usepackage{amsmath}
\usepackage{bm}
\usepackage{xcolor}
\usepackage{extarrows}
\usepackage{mathrsfs}
\usepackage{graphicx}
\usepackage{enumitem}
\usepackage{multirow}
\usepackage{booktabs}

\ifCLASSOPTIONcompsoc
  \usepackage[nocompress]{cite}
\else
  \usepackage{cite}
\fi
\ifCLASSINFOpdf
\else
\fi

\makeatletter
\def\mydownright{$\m@th\ \ \setbox\z@\hbox{$.$}\searrow\leaders\hbox{\raisebox{-0.18em}{\vrule \@depth0.05em}} \hfill\nearrow\quad\!$}
\def\mydownrightnarrow{$\m@th\quad\ \,\setbox\z@\hbox{$.$}\searrow\leaders\hbox{\raisebox{-0.18em}{\vrule \@depth0.05em}} \hfill\nearrow\quad\qquad$}
\def\myupright{$\m@th\ \ \setbox\z@\hbox{$.$}\nearrow\leaders\hbox{\raisebox{0.72em}{\vrule \@depth0.05em}} \hfill\searrow\quad\!$}
\def\mydoubleupright{$\m@th\ \ \setbox\z@\hbox{$.$}\nearrow\leaders\hbox{\raisebox{0.72em}{\vrule \@depth0.05em}} \hfill\!\leaders\hbox{\raisebox{0.72em}{\vrule \@depth0.05em}}\hfill\searrow\leaders\hbox{\raisebox{0.72em}{\vrule \@depth0.05em}}\hfill\searrow\quad\!$}

\def\myunderright#1{\mathop{\vtop{\m@th\ialign{##\crcr
				$\hfil\displaystyle{#1}\hfil$\crcr
				\noalign{\kern3\p@\nointerlineskip}%
				\mydownright\crcr\noalign{\kern3\p@}}}}\limits}
\def\myunderrightnarrow#1{\mathop{\vtop{\m@th\ialign{##\crcr
				$\hfil\displaystyle{#1}\hfil$\crcr
				\noalign{\kern3\p@\nointerlineskip}%
				\mydownrightnarrow\crcr\noalign{\kern3\p@}}}}\limits}
\def\myoverright#1{\mathop{\vbox{\m@th\ialign{##\crcr\noalign{\kern3\p@}%
				\myupright\crcr\noalign{\kern3\p@\nointerlineskip}%
				$\hfil\displaystyle{#1}\hfil$\crcr}}}\limits}
\def\mydoubleright#1{\mathop{\vbox{\m@th\ialign{##\crcr\noalign{\kern3\p@}%
				\mydoubleupright\crcr\noalign{\kern3\p@\nointerlineskip}%
				$\hfil\displaystyle{#1}\hfil$\crcr}}}\limits}
\makeatother
\newcommand\myfrac{\genfrac{}{}{0pt}0}
\DeclareMathAlphabet{\mathpzc}{OT1}{pzc}{m}{it}

\newtheorem{definition}{Definition}

\begin{document}
\title{A Comprehensive Survey on Transfer Learning}

\author{Fuzhen Zhuang, Zhiyuan Qi, Keyu Duan, Dongbo Xi, Yongchun Zhu, Hengshu Zhu,~\IEEEmembership{Senior Member,~IEEE,} Hui Xiong,~\IEEEmembership{Fellow,~IEEE,} and Qing He

\IEEEcompsocitemizethanks{\IEEEcompsocthanksitem Fuzhen Zhuang, Zhiyuan Qi, Keyu Duan, Dongbo Xi, Yongchun Zhu, and Qing He are with the Key Laboratory of Intelligent Information Processing of Chinese Academy of Sciences (CAS), Institute of Computing Technology, CAS, Beijing 100190, China and the University of Chinese Academy of Sciences, Beijing 100049, China.
\IEEEcompsocthanksitem Hengshu Zhu is with Baidu Inc., No. 10 Shangdi 10th Street, Haidian District, Beijing, China.	
\IEEEcompsocthanksitem Hui Xiong is with Rutgers, the State University of New Jersey, 1 Washington Park, Newark, New Jersey, USA.
\IEEEcompsocthanksitem Zhiyuan Qi is with the equal contribution to the first author.
\IEEEcompsocthanksitem Fuzhen Zhuang and Zhiyuan Qi are corresponding authors, zhuangfuzhen@ict.ac.cn and qizhyuan@gmail.com.
}

\markboth{A Comprehensive Survey on Transfer Learning}}%

\IEEEtitleabstractindextext{%
\begin{abstract}
Transfer learning aims at improving the performance of target learners on target domains by transferring the knowledge contained in different but related source domains. In this way, the dependence on a large number of target domain data can be reduced for constructing target learners. Due to the wide application prospects, transfer learning has become a popular and promising area in machine learning. Although there are already some valuable and impressive surveys on transfer learning, these surveys introduce approaches in a relatively isolated way and lack the recent advances in transfer learning. Due to the rapid expansion of the transfer learning area, it is both necessary and challenging to comprehensively review the relevant studies. This survey attempts to connect and systematize the existing transfer learning researches, as well as to summarize and interpret the mechanisms and the strategies of transfer learning in a comprehensive way, which may help readers have a better understanding of the current research status and ideas. Unlike previous surveys, this survey paper reviews more than forty representative transfer learning approaches, especially homogeneous transfer learning approaches, from the perspectives of data and model. The applications of transfer learning are also briefly introduced. In order to show the performance of different transfer learning models, over twenty representative transfer learning models are used for experiments. The models are performed on three different datasets, i.e., Amazon Reviews, Reuters-21578, and Office-31. And the experimental results demonstrate the importance of selecting appropriate transfer learning models for different applications in practice.
\end{abstract}     

\begin{IEEEkeywords}
Transfer learning, machine learning, domain adaptation, interpretation.
\end{IEEEkeywords}}

\maketitle

\IEEEdisplaynontitleabstractindextext

\IEEEpeerreviewmaketitle

\ifCLASSOPTIONcompsoc
\IEEEraisesectionheading{\section{Introduction}}
\else
\section{Introduction}
\fi
\IEEEPARstart{A}{lthough} traditional machine learning technology has achieved great success and has been successfully applied in many practical applications, it still has some limitations for certain real-world scenarios. The ideal scenario of machine learning is that there are abundant labeled training instances, which have the same distribution as the test data. However, collecting sufficient training data is often expensive, time-consuming, or even unrealistic in many scenarios. Semi-supervised learning can partly solve this problem by relaxing the need of mass labeled data. Typically, a semi-supervised approach only requires a limited number of labeled data, and it utilizes a large amount of unlabeled data to improve the learning accuracy. But in many cases, unlabeled instances are also difficult to collect, which usually makes the resultant traditional models unsatisfactory.

\begin{figure}
	\centering
	\includegraphics[width=0.95\linewidth]{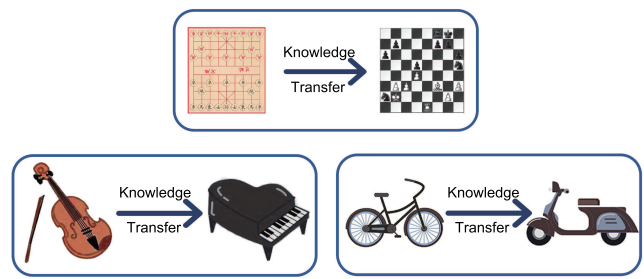}
	\caption{Intuitive examples about transfer learning.}
	\label{fig:examples}
\end{figure}
Transfer learning, which focuses on transferring the knowledge across domains, is a promising machine learning methodology for solving the above problem. The concept about transfer learning may initially come from educational psychology. According to the generalization theory of transfer, as proposed by psychologist C.H. Judd, learning to transfer is the result of the generalization of experience. It is possible to realize the transfer from one situation to another, as long as a person generalizes his experience. According to this theory, the prerequisite of transfer is that there needs to be a connection between two learning activities. In practice, a person who has learned the violin can learn the piano faster than others, since both the violin and the piano are musical instruments and may share some common knowledge. Fig. \ref{fig:examples} shows some intuitive examples about transfer learning. Inspired by human beings' capabilities to transfer knowledge across domains, transfer learning aims to leverage knowledge from a related domain (called source domain) to improve the learning performance or minimize the number of labeled examples required in a target domain. It is worth mentioning that the transferred knowledge does not always bring a positive impact on new tasks. If there is little in common between domains, knowledge transfer could be unsuccessful. For example, learning to ride a bicycle cannot help us learn to play the piano faster. Besides, the similarities between domains do not always facilitate learning, because sometimes the similarities may be misleading. For example, although Spanish and French have a close relationship with each other and both belong to the Romance group of languages, people who learn Spanish may experience difficulties in learning French, such as using the wrong vocabulary or conjugation. This occurs because previous successful experience in Spanish can interfere with learning the word formation, usage, pronunciation, conjugation, etc., in French. In the field of psychology, the phenomenon that previous experience has a negative effect on learning new tasks is called negative transfer \cite{PS1992OEP}. Similarly, in the transfer learning area, if the target learner is negatively affected by the transferred knowledge, the phenomenon is also termed as negative transfer \cite{PY2010TKDE, WDP2019CVPR}. Whether negative transfer will occur may depend on several factors, such as the relevance between the source and the target domains and the learner’s capacity of finding the transferable and beneficial part of the knowledge across domains. In \cite{WDP2019CVPR}, a formal definition and some analyses of negative transfer are given.

Roughly speaking, according to the discrepancy between domains, transfer learning can be further divided into two categories, i.e., homogeneous and heterogeneous transfer learning \cite{WKW2016JBD}. Homogeneous transfer learning approaches are developed and proposed for handling the situations where the domains are of the same feature space. In homogeneous transfer learning, some studies assume that domains differ only in marginal distributions. Therefore, they adapt the domains by correcting the sample selection bias \cite{HSG2006NIPS} or covariate shift \cite{SSN2008AISM}. However, this assumption does not hold in many cases. For example, in sentiment classification problem, a word may have different meaning tendencies in different domains. This phenomenon is also called context feature bias \cite{DK2017JBD}. To solve this problem, some studies further adapt the conditional distributions. Heterogeneous transfer learning refers to the knowledge transfer process in the situations where the domains have different feature spaces. In addition to distribution adaptation, heterogeneous transfer learning requires feature space adaptation \cite{DK2017JBD}, which makes it more complicated than homogeneous transfer learning.

The survey aims to give readers a comprehensive understanding about transfer learning from the perspectives of data and model. The mechanisms and the strategies of transfer learning approaches are introduced to allow readers grasp how the approaches work. And a number of the existing transfer learning researches are connected and systematized. Specifically, over forty representative transfer learning approaches are introduced. Besides, we conduct experiments to demonstrate on which dataset a transfer learning model performs well.

In this survey, we focus more on homogeneous transfer learning. Some interesting transfer learning topics are not covered in this survey, such as reinforcement transfer learning \cite{TS2009JMLR}, lifelong transfer learning \cite{AEL2015IJCAI}, and online transfer learning \cite{ZH2010ICML}. The rest of this survey are organized into seven sections. Section \ref{SEC.RW} clarifies the difference between transfer learning and other related machine learning techniques. Section \ref{SEC.OV} introduces the notations used in this survey and the definitions about transfer learning. Sections \ref{SEC.DBI} and \ref{SEC.MBI} interpret transfer learning approaches from the data and the model perspectives, respectively. Section \ref{SEC.AC} introduces some applications of transfer learning. Experiments are conducted and the results are provided in Section \ref{SEC.EX}. The last section concludes this survey. The main contributions of this survey are summarized below.

\begin{itemize}
\item Over forty representative transfer learning approaches are introduced and summarized, which can give readers a comprehensive overview about transfer learning.

\item Experiments are conducted to compare different transfer learning approaches. The performance of over twenty different approaches is displayed intuitively and then analyzed, which may be instructive for the readers to select the appropriate ones in practice.
\end{itemize}

\section{Related Work} \label{SEC.RW}
Some areas related to transfer learning are introduced. The connections and difference between them and transfer learning are clarified.

\noindent{\textbf{Semi-Supervised Learning} \cite{CSZ2010MIT}:} Semi-supervised learning is a machine learning task and method that lies between supervised learning (with completely labeled instances) and unsupervised learning (without any labeled instances). Typically, a semi-supervised method utilizes abundant unlabeled instances combined with a limited number of labeled instances to train a learner. Semi-supervised learning relaxes the dependence on labeled instances, and thus reduces the expensive labeling costs. Note that, in semi-supervised learning, both the labeled and unlabeled instances are drawn from the same distribution. In contrast, in transfer learning, the data distributions of the source and the target domains are usually different. Many transfer learning approaches absorb the technology of semi-supervised learning. The key assumptions in semi-supervised learning, i.e., smoothness, cluster, and manifold assumptions, are also made use of in transfer learning. It is worth mentioning that semi-supervised transfer learning is a controversial term. The reason is that the concept of whether the label information is available in transfer learning is ambiguous because both the source and the target domains can be involved.

\noindent{\textbf{Multi-View Learning} \cite{S2013NCA}:} Multi-view learning focuses on the machine learning problems with multi-view data. A view represents a distinct feature set. An intuitive example about multiple views is that a video object can be described from two different viewpoints, i.e., the image signal and the audio signal. Briefly, multi-view learning describes an object from multiple views, which results in abundant information. By properly considering the information from all views, the learner's performance can be improved. There are several strategies adopted in multi-view learning such as subspace learning, multi-kernel learning, and co-training \cite{XTX2013ARXIV,ZXX2017IF}. Multi-view techniques are also adopted in some transfer learning approaches. For example, Zhang {\it et al.} proposed a multi-view transfer learning framework, which imposes the consistency among multiple views \cite{ZHL2011KDD}. Yang and Gao incorporated multi-view information across different domains for knowledge transfer \cite{YG2013IJCAI}. The work by Feuz and Cook introduces a multi-view transfer learning approach for activity learning, which transfers activity knowledge between heterogeneous sensor platforms \cite{FC2017KIS}.

\noindent{\textbf{Multi-Task Learning} \cite{ZY2018NSR}:} The thought of multi-task learning is to jointly learn a group of related tasks. To be more specific, multi-task learning reinforces each task by taking advantage of the interconnections between task, i.e., considering both the inter-task relevance and the inter-task difference. In this way, the generalization of each task is enhanced. The main difference between transfer learning and multi-task learning is that the former transfer the knowledge contained in the related domains, while the latter transfer the knowledge via simultaneously learning some related tasks. In other words, multi-task learning pays equal attention to each task, while transfer learning pays more attention to the target task than to the source task. There are some commons and associations between transfer learning and multi-task learning. Both of them aim to improve the performance of learners via knowledge transfer. Besides, they adopt some similar strategies for constructing models, such as feature transformation and parameter sharing. Note that some existing studies utilize both the transfer learning and the multi-task learning technologies. For example, the work by Zhang {\it et al.} employs multi-task and transfer learning techniques for biological image analysis \cite{ZLZ2015KDD}. The work by Liu {\it et al.} proposes a framework for human action recognition based on multi-task learning and multi-source transfer learning \cite{LXN2019TIP}.

\section{Overview} \label{SEC.OV}
\begin{figure*}
	\centering
	\includegraphics[width=1.5\columnwidth]{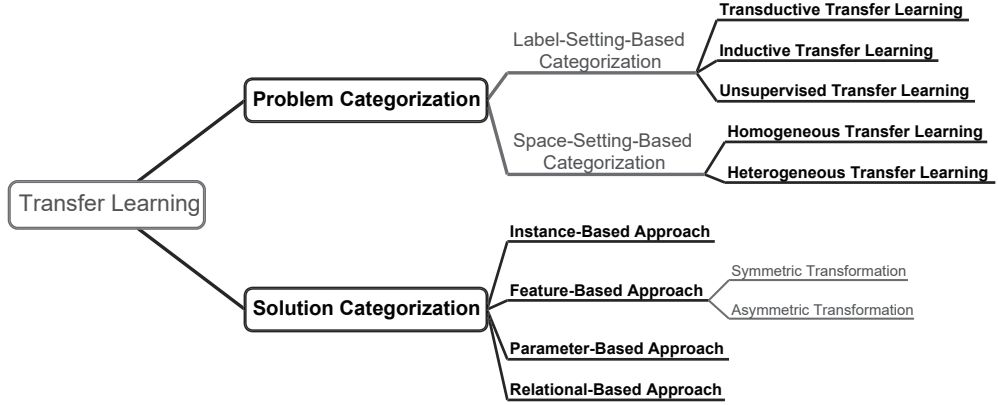}
	\caption{Categorizations of transfer learning.}
	\label{Fig1}
\end{figure*}
In this section, the notations used in this survey are listed for convenience.
Besides, some definitions and categorizations about transfer learning are introduced, and some related surveys are also provided.
\subsection{Notation}
\begin{table}[!ht]
\caption{Notations.}
\label{table.notation}
\begin{center}
	\begin{tabular}[0.98]{l l}\hline
		Symbol  & Definition \\
		\hline
		$n$ & Number of instances \\
		$m$ & Number of domains \\
		$\mathcal{D}$ & Domain \\
		$\mathcal{T}$ & Task \\
		$\mathcal{X}$ & Feature space \\
		$\mathcal{Y}$ & Label space \\
		$\mathbf{x}$ & Feature vector \\
		$y$ & Label \\
		$X$ & Instance set \\
		$Y$ & Label set corresponding to $X$\\
		$S$  & Source domain \\
		$T$  & Target domain  \\
		$L$  & Labeled instances \\
		$U$  & Unlabeled instances \\
		$\mathcal{H}$ & Reproducing kernel Hilbert space \\
		$\bm\theta$ & Mapping/Coefficient vector \\
		$\alpha$ & Weighting coefficient \\
		$\beta$ & Weighting coefficient \\
		$\lambda$ & Tradeoff parameter \\
		$\delta$ & Parameter/Error \\
		$b$ & Bias \\
		$B$ & Boundary parameter \\	
		$N$ & Iteration/Kernel number\\
		$f$  & Decision function  \\
		$\mathcal{L}$ & Loss function \\
		$\eta$  & Scale parameter\\	
		$G$ & Graph\\
		$\Phi$  & Nonlinear mapping \\
		$\sigma$ & Monotonically increasing function\\
		$\Omega$ & Structural risk\\
		$\kappa$  & Kernel function \\
		$K$ & Kernel matrix\\
		$H$ & Centering matrix\\
		$C$ & Covariance matrix\\
		$d$ & Document\\
		$w$ & Word\\
		$z$ & Class variable\\
		$\tilde{\mathbf{z}}$ & Noise\\
		$\mathpzc{D}$ & Discriminator\\
		$\mathpzc{G}$ & Generator\\
		$S$ & Function\\
		$M$ & Orthonormal bases\\
		$\Theta$ & Model parameters\\
		$P$  & Probability \\
		$\mathbb{E}$ & Expectation \\
		$Q$ & Matrix variable\\
		$R$ & Matrix variable\\
		$W$ & Mapping matrix \\
		\hline
	\end{tabular}
\end{center}
\end{table}
For convenience, a list of symbols and their definitions are shown in Table \ref{table.notation}. Besides, we use $||\cdot||$ to represent the norm and superscript $^\text{T}$ to denote the transpose of a vector/matrix.
\subsection{Definition}
In this subsection, some definitions about transfer learning are given. Before giving the definition of transfer learning, let us review the definitions of a domain and a task.
\begin{definition}
(Domain) {\it A domain $\mathcal{D}$ is composed of two parts, i.e., a feature space $\mathcal{X}$ and a marginal distribution $P(X)$. In other words, $\mathcal{D}=\{\mathcal{X},P(X)\}$. And the symbol $X$ denotes an instance set, which is defined as $X=\{\mathbf{x}|\mathbf{x}_i\in\mathcal{X},~i=1,\cdots,n\}$.}
\end{definition}

\begin{definition}
(Task)
{\it A task $\mathcal{T}$ consists of a label space $\mathcal{Y}$ and a decision function $f$, i.e., $\mathcal{T}=\{\mathcal{Y},f\}$. The decision function $f$ is an implicit one, which is expected to be learned from the sample data.}
\end{definition}
Some machine learning models actually output the predicted conditional distributions of instances. In this case, $f(\mathbf{x}_{j})=\{P(y_{k}|\mathbf{x}_{j})|y_{k}\in\mathcal{Y},k=1,\cdots,|\mathcal{Y}|\}$.

In practice, a domain is often observed by a number of instances with/without the label information. For example, a source domain $\mathcal{D}_{S}$ corresponding to a source task $\mathcal{T}_{S}$ is usually observed via the instance-label pairs, i.e., $D_{S}=\{(\mathbf{x},y)|\mathbf{x}_{i}\in\mathcal{X}^{S},y_{i}\in\mathcal{Y}^{S},i=1,\cdots,n^{S}\}$; an observation of the target domain usually consists of a number of unlabeled instances and/or limited number of labeled instances.
\begin{definition}
(Transfer Learning)
{\it Given some/an observation(s) corresponding to $m^{S}\in\mathbb{N}^{+}$ source domain(s) and task(s) (i.e., $\{(\mathcal{D}_{S_{i}},\mathcal{T}_{S_{i}})|i=1,\cdots,m^{S}\}$), and some/an observation(s) about $m^{T}\in\mathbb{N}^{+}$ target domain(s) and task(s) (i.e., $\{(\mathcal{D}_{T_{j}},\mathcal{T}_{T_{j}})|j=1,\cdots,m^{T}\}$), transfer learning utilizes the knowledge implied in the source domain(s) to improve the performance of the learned decision functions $f^{T_j}$ ($j=1,\cdots,m^{T}$) on the target domain(s).}
\end{definition}
The above definition, which covers the situation of multi-source transfer learning, is an extension of the one presented in the survey \cite{PY2010TKDE}. If $m^{S}$ equals $1$, the scenario is called single-source transfer learning. Otherwise, it is called multi-source transfer learning. Besides, $m^{T}$ represents the number of the transfer learning tasks. A few studies focus on the setting that $m^{T}\ge 2$ \cite{PHS2019ICML}. The existing transfer learning studies pay more attention to the scenarios where $m^{T}=1$ (especially where $m^{S}=m^{T}=1$). It is worth mentioning that the observation of a domain or a task is a concept with broad sense, which is often cemented into a labeled/unlabeled instance set or a pre-learned model. A common scenario is that we have abundant labeled instances or have a well-trained model on the source domain, and we only have limited labeled target-domain instances. In this case, the resources such as the instances and the model are actually the observations, and the goal of transfer learning is to learn a more accurate decision function on the target domain.

Another term commonly used in the transfer learning area is domain adaptation. Domain adaptation refers to the process that adapting one or more source domains to transfer knowledge and improve the performance of the target learner \cite{WKW2016JBD}. Transfer learning often relies on the domain adaptation process, which attempts to reduce the difference between domains.

\subsection{Categorization of Transfer Learning}
There are several categorization criteria of transfer learning. For example, transfer learning problems can be divided into three categories, i.e., transductive, inductive, and unsupervised transfer learning \cite{PY2010TKDE}. The complete definitions of these three categories are presented in \cite{PY2010TKDE}. These three categories can be interpreted from a label-setting aspect. Roughly speaking, transductive transfer learning refers to the situations where the label information only comes from the source domain. If the label information of the target-domain instances is available, the scenario can be categorized into inductive transfer learning. If the label information is unknown for both the source and the target domains, the situation is known as unsupervised transfer learning. Another categorization is based on the consistency between the source and the target feature spaces and label spaces. If $\mathcal{X}^{S}=\mathcal{X}^{T}$ and $\mathcal{Y}^{S}=\mathcal{Y}^{T}$, the scenario is termed as homogeneous transfer learning. Otherwise, if $\mathcal{X}^{S}\ne\mathcal{X}^{T}$ or/and $\mathcal{Y}^{S}\ne\mathcal{Y}^{T}$, the scenario is referred to as heterogeneous transfer learning.

According to the survey \cite{PY2010TKDE}, the transfer learning approaches can be categorized into four groups: instance-based, feature-based, parameter-based, and relational-based approaches. Instance-based transfer learning approaches are mainly based on the instance weighting strategy. Feature-based approaches transform the original features to create a new feature representation; they can be further divided into two subcategories, i.e., asymmetric and symmetric feature-based transfer learning. Asymmetric approaches transform the source features to match the target ones. In contrast, symmetric approaches attempt to find a common latent feature space and then transform both the source and the target features into a new feature representation. The parameter-based transfer learning approaches transfer the knowledge at the model/parameter level. Relational-based transfer learning approaches mainly focus on the problems in relational domains. Such approaches transfer the logical relationship or rules learned in the source domain to the target domain. For better understanding, Fig. \ref{Fig1} shows the above-mentioned categorizations of transfer learning.

Some surveys are provided for the readers who want to have a more complete understanding of this field. The survey by Pan and Yang \cite{PY2010TKDE}, which is a pioneering work, categorizes transfer learning and reviews the research progress before $2010$. The survey by Weiss {\it et al.} introduces and summarizes a number of homogeneous and heterogeneous transfer learning approaches \cite{WKW2016JBD}. Heterogeneous transfer learning is specially reviewed in the survey by Day and Khoshgoftaar \cite{DK2017JBD}. Some surveys review the literatures related to specific themes such as reinforcement learning \cite{TS2009JMLR}, computational intelligence \cite{LBH2015KBS}, and deep learning \cite{TSK2018ICANN,WD2018NC}. Besides, a number of surveys focus on specific application scenarios including activity recognition \cite{CFK2013KIS}, visual categorization \cite{SZL2015TNNLS}, collaborative recommendation \cite{P2016NC}, computer vision \cite{WD2018NC}, and sentiment analysis \cite{LSJ2019ACCESS}.

Note that the organization of this survey does not strictly follow the above-mentioned categorizations. In the next two sections, transfer learning approaches are interpreted from the data and the model perspectives. Roughly speaking, data-based interpretation covers the above-mentioned instance-based and feature-based transfer learning approaches, but from a broader perspective. Model-based interpretation covers the above-mentioned parameter-based approaches. Since there are relatively few studies concerning relational-based transfer learning and the representative approaches are well introduced in \cite{PY2010TKDE,WKW2016JBD}, this survey does not focus on relational-based approaches.

\section{Data-Based Interpretation} \label{SEC.DBI}
\begin{figure*}
	\centering
	\includegraphics[width=1.5\columnwidth]{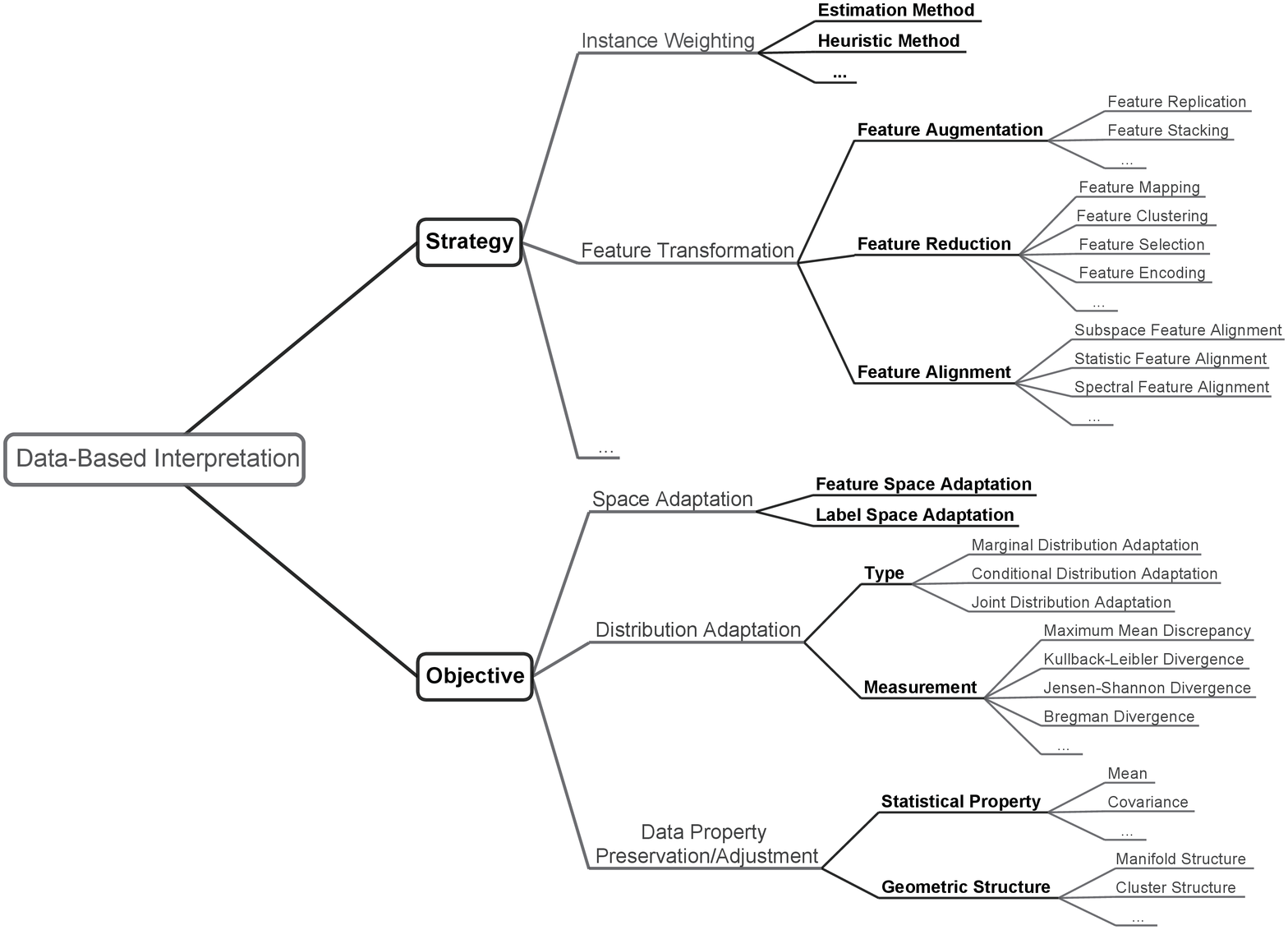}
	\caption{Strategies and the objectives of the transfer learning approaches from the data perspective.}
	\label{Fig2}
\end{figure*}
Many transfer learning approaches, especially the data-based approaches, focus on transferring the knowledge via the adjustment and the transformation of data. Fig. \ref{Fig2} shows the strategies and the objectives of the approaches from the data perspective. As shown in Fig. \ref{Fig2}, space adaptation is one of the objectives. This objective is required to be satisfied mostly in heterogeneous transfer learning scenarios. In this survey, we focus more on homogeneous transfer learning, and the main objective in this scenario is to reduce the distribution difference between the source-domain and the target-domain instances. Furthermore, some advanced approaches may attempt to preserve the data properties in the adaptation process. There are generally two strategies for realizing the objectives from the data perspective, i.e., instance weighting and feature transformation. In this section, some related transfer learning approaches are introduced in proper order according to the strategies shown in Fig. \ref{Fig2}.
\subsection{Instance Weighting Strategy} \label{SEC.IWS}
Let us first consider a simple scenario in which a large number of labeled source-domain and a limited number of target-domain instances are available and domains differ only in marginal distributions (i.e., $P^{S}(X)\ne P^{T}(X)$ and $P^{S}(Y|X)=P^{T}(Y|X)$). For example, suppose we need to build a model to diagnose cancer in a specific region where the elderly are the majority. Limited target-domain instances are given, and relevant data are available from another region where young people are the majority. Directly transferring all the data from another region may be unsuccessful, since the marginal distribution difference exists, and the elderly have a higher risk of cancer than younger people.
In this scenario, it is natural to consider adapting the marginal distributions. A simple idea is to assign weights to the source-domain instances in the loss function. The weighting strategy is based on the following equation \cite{HSG2006NIPS}:
\begin{equation*}
\begin{aligned}
	\mathbb{E}_{(\mathbf{x},y)\sim P^{T}}\left[\mathcal{L}(\mathbf{x},y;f)\right]&=\mathbb{E}_{(\mathbf{x},y)\sim P^{S}}\left[\dfrac{P^{T}(\mathbf{x},y)}{P^{S}(\mathbf{x},y)}\mathcal{L}(\mathbf{x},y;f)\right]\\&=\mathbb{E}_{(\mathbf{x},y)\sim P^{S}}\left[\dfrac{P^{T}(\mathbf{x})}{P^{S}(\mathbf{x})}\mathcal{L}(\mathbf{x},y;f)\right].
\end{aligned}
\end{equation*}
Therefore, the general objective function of a learning task can be written as \cite{HSG2006NIPS}:
\begin{equation*}
{\min_{f}}{\dfrac{1}{n^{{S}}}\sum_{i=1}^{n^{{S}}}{\beta_{i}\mathcal{L}\left(f(\mathbf{x}^{S}_i),y^{S}_i\right)}+\Omega(f)},
\end{equation*}
where $\beta_{i}$ ($i=1,2,\cdots,n^{S}$) is the weighting parameter. The theoretical value of $\beta_{i}$ is equal to $P^{T}(\mathbf{x}_i)/P^{S}(\mathbf{x}_i)$. However, this ratio is generally unknown and is difficult to be obtained by using the traditional methods.

Kernel Mean Matching (KMM) \cite{HSG2006NIPS}, which is proposed by Huang {\it et al.}, resolves the estimation problem of the above unknown ratios by matching the means between the source-domain and the target-domain instances in a Reproducing Kernel Hilbert Space (RKHS), i.e.,
\begin{align*}
&\mathop{\arg\min}_{\beta_{i}\in[0,B]}{\left\vert\left\vert\dfrac{1}{n^{S}}\sum_{i=1}^{n^{S}}\beta_{i}\Phi(\mathbf{x}^{S}_{i})-\dfrac{1}{n^{T}}\sum_{j=1}^{n^{T}}\Phi(\mathbf{x}^{T}_{j})\right\vert\right\vert}^{2}_{\mathcal{H}}\\
&s.t.~\vert\dfrac{1}{n^{S}}\sum_{i=1}^{n^{S}}\beta_{i}-1\vert\leq \delta,
\end{align*}
where $\delta$ is a small parameter, and $B$ is a parameter for constraint. The above optimization problem can be converted into a quadratic programming problem by expanding and using the kernel trick. This approach to estimating the ratios of distributions can be easily incorporated into many existing algorithms. Once the weight $\beta_{i}$ is obtained, a learner can be trained on the weighted source-domain instances.

There are some other studies attempting to estimate the weights. For example, Sugiyama {\it et al.} proposed an approach termed Kullback-Leibler Importance Estimation Procedure (KLIEP) \cite{SSN2008AISM}. KLIEP depends on the minimization of the Kullback-Leibler (KL) divergence and incorporates a built-in model selection procedure. Based on the studies of weight estimation, some instance-based transfer learning frameworks or algorithms are proposed. For example, Sun {\it et al.} proposed a multi-source framework termed 2-Stage Weighting Framework for Multi-Source Domain Adaptation (2SW-MDA) with the following two stages \cite{SCP2011NIPS}.
\begin{itemize}
\item[1.] {\it Instance Weighting}: The source-domain instances are assigned with weights to reduce the marginal distribution difference, which is similar to KMM.
\item[2.] {\it Domain Weighting}: Weights are assigned to each source domain for reducing the conditional distribution difference based on the smoothness assumption \cite{BNS2006JMLR}.
\end{itemize}
Then, the source-domain instances are reweighted based on the instance weights and the domain weights. These reweighted instances and the labeled target-domain instances are used to train the target classifier.

In addition to directly estimating the weighting parameters, adjusting weights iteratively is also effective. The key is to design a mechanism to decrease the weights of the instances that have negative effects on the target learner. A representative work is TrAdaBoost \cite{DYX2007ICML}, which is a framework proposed by Dai {\it et al}. This framework is an extension of AdaBoost \cite{FS1997JCSS}. AdaBoost is an effective boosting algorithm designed for traditional machine learning tasks. In each iteration of AdaBoost, a learner is trained on the instances with updated weights, which results in a weak classifier. The weighting mechanism of instances ensures that the instances with incorrect classification are given more attention. Finally, the resultant weak classifiers are combined to form a strong classifier. TrAdaBoost extends the AdaBoost to the transfer learning scenario; a new weighting mechanism is designed to reduce the impact of the distribution difference. Specifically, in TrAdaBoost, the labeled source-domain and labeled target-domain instances are combined as a whole, i.e., a training set, to train the weak classifier. The weighting operations are different for the source-domain and the target-domain instances. In each iteration, a temporary variable $\bar{\delta}$, which measures the classification error rate on the labeled target-domain instances, is calculated. Then, the weights of the target-domain instances are updated based on $\bar{\delta}$ and the individual classification results, while the weights of the source-domain instances are updated based on a designed constant and the individual classification results. For better understanding, the formulas used in the $k$-th iteration ($k=1,\cdots,N$) for updating the weights are presented repeatedly as follows \cite{DYX2007ICML}:
\begin{equation*}
\begin{aligned}
	&\beta^{S}_{k,i}=\beta^{S}_{k-1,i}({1+\sqrt{2\ln n^{S}/N}})^{-\left\vert f_{k}(\mathbf{x}^{S}_{i})-y^{S}_{i}\right\vert}~(i=1,\cdots,n^{S}),\\
	&\beta^{T}_{k,j}=\beta^{T}_{k-1,j}({\bar{\delta}_k}/({1-\bar{\delta}_k}))^{-\left\vert f_{k}(\mathbf{x}^{T}_{j})-y^{T}_{j}\right\vert}~(j=1,\cdots,n^{T}).
\end{aligned}
\end{equation*}
Note that each iteration forms a new weak classifier. The final classifier is constructed by combining and ensembling half the number of the newly resultant weak classifiers through voting scheme.

Some studies further extend TrAdaBoost. The work by Yao and Doretto \cite{YD2010CVPR} proposes a Multi-Source TrAdaBoost (MsTrAdaBoost) algorithm, which mainly has the following two steps in each iteration.
\begin{itemize}
\item[1.] {\it Candidate Classifier Construction}: A group of candidate weak classifiers are respectively trained on the weighted instances in the pairs of each source domain and the target domain, i.e., $\mathcal{D}_{S_i}\cup\mathcal{D}_{T}$ ($i=1,\cdots,m^{S}$).
\item[2.] {\it Instance Weighting}: A classifier (denoted by $j$ and trained on $\mathcal{D}_{S_j}\cup\mathcal{D}_{T}$) which has the minimal classification error rate $\bar{\delta}$ on the target domain instances is selected, and then is used for updating the weights of the instances in $\mathcal{D}_{S_j}$ and $\mathcal{D}_{T}$.
\end{itemize}
Finally, the selected classifiers from each iteration are combined to form the final classifier. Another parameter-based algorithm, i.e., TaskTrAdaBoost, is also proposed in the work \cite{YD2010CVPR}, which is introduced in Section \ref{SEC.MES}.

Some approaches realize instance weighting strategy in a heuristic way. For example, Jiang and Zhai proposed a general weighting framework \cite{JZ2007ACL}. There are three terms in the framework's objective function, which are designed to minimize the cross-entropy loss of three types of instances. The following types of instances are used to construct the target classifier.
\begin{itemize}[leftmargin=*]
\item {\it Labeled Target-domain Instance}: The classifier should minimize the cross-entropy loss on them, which is actually a standard supervised learning task.
\item {\it Unlabeled Target-domain Instance}: These instances' true conditional distributions $P(y|\mathbf{x}_{i}^{T,U})$ are unknown and should be estimated. A possible solution is to train an auxiliary classifier on the labeled source-domain and target-domain instances to help estimate the conditional distributions or assign pseudo labels to these instances.
\item {\it Labeled Source-domain Instance}: The authors define the weight of $\mathbf{x}^{S,L}_{i}$ as the product of two parts, i.e., $\alpha_{i}$ and $\beta_{i}$. The weight $\beta_{i}$ is ideally equal to $P^{T}(\mathbf{x}_i)/P^{S}(\mathbf{x}_i)$, which can be estimated by non-parametric methods such as KMM or can be set uniformly in the worst case. The weight $\alpha_{i}$ is used to filter out the source-domain instances that differ greatly from the target domain.
\end{itemize}
A heuristic method can be used to produce the value of $\alpha_{i}$, which contains the following three steps.
\begin{itemize}
\item[1.] {\it Auxiliary Classifier Construction}: An auxiliary classifier trained on the labeled target-domain instances are used to classify the unlabeled source-domain instances.
\item[2.] {\it Instance Ranking}: The source-domain instances are ranked based on the probabilistic prediction results.
\item[3.] {\it Heuristic Weighting ($\beta_{i}$)}: The weights of the top-$k$ source-domain instances with wrong predictions are set to zero, and the weights of others are set to one.
\end{itemize}

\subsection{Feature Transformation Strategy}
Feature transformation strategy is often adopted in feature-based approaches. For example, consider a cross-domain text classification problem. The task is to construct a target classifier by using the labeled text data from a related domain. In this scenario, a feasible solution is to find the common latent features (e.g., latent topics) through feature transformation and use them as a bridge to transfer knowledge. Feature-based approaches transform each original feature into a new feature representation for knowledge transfer. The objectives of constructing a new feature representation include minimizing the marginal and the conditional distribution difference, preserving the properties or the potential structures of the data, and finding the correspondence between features. The operations of feature transformation can be divided into three types, i.e., feature augmentation, feature reduction, and feature alignment. Besides, feature reduction can be further divided into several types such as feature mapping, feature clustering, feature selection, and feature encoding. A complete feature transformation process designed in an algorithm may consist of several operations.

\subsubsection{Distribution Difference Metric}
One primary objective of feature transformation is to reduce the distribution difference of the source and the target domain instances. Therefore, how to measure the distribution difference or the similarity between domains effectively is an important issue.

The measurement termed Maximum Mean Discrepancy (MMD) is widely used in the field of transfer learning, which is formulated as follows \cite{BGR2006BIO}:
\begin{equation*}
\text{ MMD}(X^S,X^T)={\left\vert\left\vert\dfrac{1}{n^{S}}\sum_{i=1}^{n^{S}}\Phi(\mathbf{x}^{S}_{i})-\dfrac{1}{n^{T}}\sum_{j=1}^{n^{T}}\Phi(\mathbf{x}^{T}_{j})\right\vert\right\vert}^{2}_{\mathcal{H}}.
\end{equation*}
MMD can be easily computed by using kernel trick. Briefly, MMD quantifies the distribution difference by calculating the distance of the mean values of the instances in a RKHS. Note that the above-mentioned KMM actually produces the weights of instances by minimizing the MMD distance between domains.
\begin{table}[!t]
\caption{Metrics Adopted in Transfer Learning.}
\label{table.measurement}
\begin{center}
	\begin{tabular}[0.98]{l l}\hline
		Measurement & Related Algorithms\\
		\hline
		Maximum Mean Discrepancy \cite{BGR2006BIO}& \cite{PTK2011TNN}\cite{GKZ2014RICAI}\cite{LWD2013ICCV}\cite{LWD2014TKDE}$\cdots$\\
		Kullback-Leibler Divergence \cite{KL1951AMS}& \cite{DXY2007KDD}\cite{DYX2008ICML}\cite{DD2009ICML}\cite{ZCL2015IJCAI}$\cdots$\\
		Jensen-Shannon Divergence \cite{DLP1997ACL}& \cite{CLT2010ICDM}\cite{DMM2016ICASSP}\cite{CCJ2017SM}\cite{GAM2019ICASSP}$\cdots$\\
		Bregman Divergence \cite{B1967CMMP}& \cite{STG2010TKDE}\cite{SLZ2016GRSL}\cite{SLZ2016NPL}\cite{SZL2019ICV}$\cdots$\\
		Hilbert-Schmidt Independence Criterion \cite{GBS2005ICALT} & \cite{PTK2011TNN}\cite{WY2011AAAI}\cite{XG2015TPAMI}\cite{YKZ2018TC}$\cdots$\\
		\hline
	\end{tabular}
\end{center}
\end{table}

Table. \ref{table.measurement} lists some commonly used metrics and the related algorithms. In addition to Table. \ref{table.measurement}, there are some other measurement criteria adopted in transfer learning, including Wasserstein distance \cite{SQZ2018AAAI,LBB2019CVPR}, central moment discrepancy \cite{ZGL2017ICLR}, etc. Some studies focus on optimizing and improving the existing measurements. Take MMD as an example. Gretton {\it et al.} proposed a multi-kernel version of MMD, i.e., MK-MMD \cite{GSS2012NIPS}, which takes advantage of multiple kernels. Besides, Yan {\it et al.} proposed a weighted version of MMD \cite{YDL2017CVPR}, which attempts to address the issue of class weight bias.

\subsubsection{Feature Augmentation}
Feature augmentation operations are widely used in feature transformation, especially in symmetric feature-based approaches. To be more specific, there are several ways to realize feature augmentation such as feature replication and feature stacking. For better understanding, we start with a simple transfer learning approach which is established based on feature replication.

The work by Daum\'e proposes a simple domain adaptation method, i.e., Feature Augmentation Method (FAM) \cite{D2007ACL}. This method transforms the original features by simple feature replication. Specifically, in single-source transfer learning scenario, the feature space is augmented to three times its original size. The new feature representation consists of general features, source-specific features, and target-specific features. Note that, for the transformed source-domain instances, their target-specific features are set to zero. Similarly, for the transformed target-domain instances, their source-specific features are set to zero. The new feature representation of FAM is presented as follows:
\begin{equation*}
\Phi_{S}(\mathbf{x}^{S}_i)=\langle\mathbf{x}^{S}_i,\mathbf{x}^{S}_i,\mathbf{0}\rangle,~\Phi_{T}(\mathbf{x}^{T}_j)=\langle\mathbf{x}^{T}_j,\mathbf{0},\mathbf{x}^{T}_j\rangle,
\end{equation*}
where $\Phi_{S}$ and $\Phi_{T}$ denote the mappings to the new feature space from the source and the target domain, respectively. The final classifier is trained on the transformed labeled instances. It is worth mentioning that this augmentation method is actually redundant. In other words, augmenting the feature space in other ways (with fewer dimensions) may be able to produce competent performance. The superiority of FAM is that its feature expansion has an elegant form, which results in some good properties such as the generalization to multi-source scenarios. An extension of FAM is proposed in \cite{DKS2010NIPS} by Daum\'e {\it et al.}, which utilizes the unlabeled instances to further facilitate the knowledge transfer process.

However, FAM may not work well in handling heterogeneous transfer learning tasks. The reason is that directly replicating features and padding zero vectors are less effective when the source and the target domains have different feature representations. To solve this problem, Li {\it et al.} proposed an approach termed Heterogeneous Feature Augmentation (HFA) \cite{DXT2012ICML,LDX2014TPAMI}. The feature representation of HFA is presented below:
\begin{equation*}
\Phi_{S}(\mathbf{x}^{S}_i)=\langle W^{S}\mathbf{x}^{S}_i,\mathbf{x}^{S}_i,\mathbf{0}^{T}\rangle,~\Phi_{T}(\mathbf{x}^{T}_j)=\langle W^{T}\mathbf{x}^{T}_j,\mathbf{0}^{S},\mathbf{x}^{T}_j\rangle,
\end{equation*}
where $W^{S}\mathbf{x}^{S}_i$ and $W^{T}\mathbf{x}^{T}_j$ have the same dimension; $\mathbf{0}^{S}$ and $\mathbf{0}^{T}$ denote the zero vectors with the dimensions of $\mathbf{x}^{S}$ and $\mathbf{x}^{T}$, respectively. HFA maps the original features into a common feature space, and then performs a feature stacking operation. The mapped features, original features, and zero elements are stacked in a particular order to produce a new feature representation.

\subsubsection{Feature Mapping} \label{SEC.FE.FR}
In the field of traditional machine learning, there are many feasible mapping-based methods of extracting features such as Principal Component Analysis (PCA) \cite{DK1996WILEY} and Kernelized-PCA (KPCA) \cite{SSM1998NC}. However, these methods mainly focus on the data variance rather than the distribution difference. In order to solve the distribution difference, some feature extraction methods are proposed for transfer learning. Let us first consider a simple scenario where there is little difference in the conditional distributions of the domains. In this case, the following simple objective function can be used to find a mapping for feature extraction:
\begin{equation*}
\min_{\Phi}{\big(\text{DIST}(X^{S},X^{T};\Phi)+\lambda\Omega(\Phi)\big)/\big(\text{VAR}(X^{S}\cup X^{T};\Phi)\big)},
\end{equation*}
where $\Phi$ is a low-dimensional mapping function, $\text{DIST}(\cdot)$ represents a distribution difference metric, $\Omega(\Phi)$ is a regularizer controlling the complexity of $\Phi$, and $\text{VAR}(\cdot)$ represents the variance of instances. This objective function aims to find a mapping function $\Phi$ that minimizes the marginal distribution difference between domains and meanwhile makes the variance of the instances as large as possible. The objective corresponding to the denominator can be optimized in several ways. One possible way is to optimize the objective of the numerator with a variance constraint. For example, the scatter matrix of the mapped instances can be enforced as an identity matrix. Another way is to optimize the objective of the numerator in a high-dimensional feature space at first. Then, a dimension reduction algorithm such as PCA or KPCA can be performed to realize the objective of the denominator.

Further, finding the explicit formulation of $\Phi(\cdot)$ is nontrivial. To solve this problem, some approaches adopt linear mapping technique or turn to the kernel trick. In general, there are three main ideas to deal with the above optimization problems.
\begin{itemize}[leftmargin=*]
\item (Mapping Learning $+$ Feature Extraction) A possible way is to find a high-dimensional space at first where the objectives are met by solving a kernel matrix learning problem or a transformation matrix finding problem. Then, the high-dimensional features are compacted to form a low-dimensional feature representation. For example, once the kernel matrix is learned, the principal components of the implicit high-dimensional features can be extracted to construct a new feature representation based on PCA.

\item (Mapping Construction $+$ Mapping Learning) Another way is to map the original features to a constructed high-dimensional feature space, and then a low-dimensional mapping is learned to satisfy the objective function. For example, a kernel matrix can be constructed based on a selected kernel function at first. Then, the transformation matrix can be learned, which projects the high-dimensional features into a common latent subspace.

\item (Direct Low-dimensional Mapping Learning) It is usually difficult to find a desired low-dimensional mapping directly. However, if the mapping is assumed to satisfy certain conditions, it may be solvable. For example, if the low-dimensional mapping is restricted to be a linear one, the optimization problem can be easily solved.
\end{itemize}

Some approaches also attempt to match the conditional distributions and preserve the structures of the data. To achieve this, the above simple objective function needs to incorporate new terms or/and constraints. For example, the following general objective function is a possible choice:
\begin{equation*}
\begin{aligned}
	&\min_{\Phi}\mu\text{DIST}(X^{S},X^{T};\Phi)+\lambda_{1}\Omega^{\text{GEO}}(\Phi)+\lambda_{2}\Omega(\Phi)\\&\qquad+(1-\mu)\text{DIST}(Y^{S}|X^{S},Y^{T}|X^{T};\Phi),\\
	& s.t.~\Phi(X)^{\text{T}}H\Phi(X)=I,~\text{with}~H=I-(\mathbf{1}/{n})\in\mathbb{R}^{n\times n},\\
\end{aligned}
\end{equation*}
where $\mu$ is a parameter balancing the marginal and the conditional distribution difference \cite{WCH2017ICDM}, $\Omega^{\text{GEO}}(\Phi)$ is a regularizer controlling the geometric structure, $\Phi(X)$ is the matrix whose rows are the instances from both the source and the target domains with the extracted new feature representation, $H$ is the centering matrix for constructing the scatter matrix, and the constraint is used to maximize the variance. The last term in the objective function denotes the measurement of the conditional distribution difference.

Before the further discussion about the above objective function, it is worth mentioning that the label information of the target-domain instances is often limited or even unknown. The lack of the label information makes it difficult to estimate the distribution difference. In order to solve this problem, some approaches resort to the pseudo-label strategy, i.e., assigning pseudo labels to the unlabeled target-domain instances. A simple method of realizing this is to train a base classifier to assign pseudo labels. By the way, there are some other methods of providing pseudo labels such as co-training \cite{BM1998ACCLT,CWB2011NIPS} and tri-training \cite{ZL2005TKDE,SUH2017ICML}. Once the pseudo-label information is complemented, the conditional distribution difference can be measured. For example, MMD can be modified and extended to measure the conditional distribution difference. Specifically, for each label, the source-domain and the target-domain instances that belong to the same class are collected, and the estimation expression of the conditional distribution difference is given by \cite{LWD2013ICCV}:
\begin{equation*}
\sum_{k=1}^{|\mathcal{Y}|}{\left\vert\left\vert\dfrac{1}{n^{S}_{k}}\sum_{i=1}^{n^{S}_{k}}\Phi(\mathbf{x}^{S}_{i})-\dfrac{1}{n^{T}_{k}}\sum_{j=1}^{n^{T}_{k}}\Phi(\mathbf{x}^{T}_{j})\right\vert\right\vert}^{2}_{\mathcal{H}},
\end{equation*}
where $n^{S}_{k}$ and $n^{T}_{k}$ denote the numbers of the instances in the source and the target domains with the same label $\mathcal{Y}_{k}$, respectively. This estimation actually measures the class-conditional distribution (i.e., $P(\mathbf{x}|y)$) difference to approximate the conditional distribution (i.e., $P(y|\mathbf{x})$) difference. Some studies improve the above estimation. For example, the work by Wang {\it et al.} uses a weighted method to additionally solve the class imbalance problem \cite{WCH2017ICDM}. For better understanding, the transfer learning approaches that are the special cases of the general objective function presented in the previous paragraph are detailed as follows.
\begin{itemize}[leftmargin=*]
\item ($\mu=1$ and $\lambda_{1}\neq 0$) The objective function of Maximum Mean Discrepancy Embedding (MMDE) is given by \cite{PKY2008AAAI}:
\begin{equation*}
	\begin{aligned}
		&\min_{K}{\text{MMD}(X^{S},X^{T};\Phi)-\dfrac{\lambda_{1}}{n^{S}+n^{T}}\sum_{i\ne j}\vert\vert\Phi(\mathbf{x}_{i})-\Phi(\mathbf{x}_j)\vert\vert^2}\\
		& s.t.~\forall(\mathbf{x}_{i}\in k\text{-NN}(\mathbf{x}_{j}))\land(\mathbf{x}_{j}\in k\text{-NN}(\mathbf{x}_{i})),\\
		&\quad~~~\vert\vert\Phi(\mathbf{x}_{i})-\Phi(\mathbf{x}_j)\vert\vert^2=\vert\vert\mathbf{x}_{i}-\mathbf{x}_{j}\vert\vert^2, (\mathbf{x}_{i},\mathbf{x}_{j}\in X^{S}\cup X^{T}),
	\end{aligned}
\end{equation*}
where $k$-NN$(\mathbf{x})$ represents the $k$ nearest neighbors of the instance $\mathbf{x}$. The authors design the above objective function motivated by Maximum Variance Unfolding (MVU) \cite{WSS2004ICML}. Instead of employing a scatter matrix constraint, the constraints and the second term of this objective function aim to maximize the distance between instances as well as preserve local geometry. The desired kernel matrix $K$ can be learned by solving a Semi-Definite Programming (SDP) \cite{VB1996SR} problem. After obtaining the kernel matrix, PCA is applied to it, and then the leading eigenvectors are selected to help construct a low-dimensional feature representation.

\item ($\mu=1$ and $\lambda_{1}=0$) The work by Pan {\it et al.} proposes an approach termed Transfer Component Analysis (TCA) \cite{PTK2009IJCAI,PTK2011TNN}. TCA adopts MMD to measure the marginal distribution difference and enforces the scatter matrix as the constraint. Different from MMDE that learns the kernel matrix and then further adopts PCA, TCA is a unified method that just needs to learn a linear mapping from an empirical kernel feature space to a low-dimensional feature space. In this way, it avoids solving the SDP problem, which results in relatively low computational burden. The final optimization problem can be easily solved via eigen-decomposition. TCA can also be extended to utilize the label information. In the extended version, the scatter matrix constraint is replaced by a new one that balances the label dependence (measured by HSIC) and the data variance. Besides, a graph Laplacian regularizer \cite{BNS2006JMLR} is also added to preserve the geometry of the manifold. Similarly, the final optimization problem can also be solved by eigen-decomposition.

\item ($\mu=0.5$ and $\lambda_{1}=0$) Long {\it et al.} proposed an approach termed Joint Distribution Adaptation (JDA) \cite{LWD2013ICCV}. JDA attempts to find a transformation matrix that maps the instances to a low-dimensional space where both the marginal and the conditional distribution difference are minimized. To realize it, the MMD metric and the pseudo-label strategy are adopted. The desired transformation matrix can be obtained by solving a trace optimization problem via eigen-decomposition. Further, it is obvious that the accuracy of the estimated pseudo labels affects the performance of JDA. In order to improve the labeling quality, the authors adopt the iterative refinement operations. Specifically, in each iteration, JDA is performed, and then a classifier is trained on the instances with the extracted features. Next, the pseudo labels are updated based on the trained classifier. After that, JDA is performed repeatedly with the updated pseudo labels. The iteration ends when convergence occurs. Note that JDA can be extended by utilizing the label and structure information \cite{HTY2016TIP}, clustering information \cite{TH2017KIS}, various statistical and geometrical information \cite{ZLO2017CVPR}, etc.

\item ($\mu\in(0,1)$ and $\lambda_{1}=0$) The paper by Wang {\it et al.} proposes an approach termed Balanced Distribution Adaptation (BDA) \cite{WCH2017ICDM}, which is an extension of JDA. Different from JDA which assumes that the marginal and the conditional distributions have the same importance in adaptation, BDA attempts to balance the marginal and the conditional distribution adaptation. The operations of BDA are similar to JDA. In addition, the authors also proposed the Weighted BDA (WBDA). In WBDA, the conditional distribution difference is measured by a weighted version of MMD to solve the class imbalance problem.
\end{itemize}

It is worth mentioning that some approaches transform the features into a new feature space (usually of a high dimension) and train an adaptive classifier simultaneously. To realize this, the mapping function of the features and the decision function of the classifier need to be associated. One possible way is to define the following decision function: $f(\mathbf{x})={\bm\theta}\cdot\Phi(\mathbf{x})+b$, where ${\bm\theta}$ denotes the classifier parameter; $b$ denotes the bias. In light of the representer theorem \cite{SHS2001COLT}, the parameter ${\bm\theta}$ can be defined as ${\bm\theta}=\sum_{i=1}^{n}{\alpha_i \Phi(\mathbf{x}_i)},$ and thus we have
\begin{equation*}
\begin{aligned}
	f(\mathbf{x})=\sum_{i=1}^{n}{\alpha_i \Phi(\mathbf{x}_i)\cdot\Phi(\mathbf{x})}+b=\sum_{i=1}^{n}{\alpha_i\kappa(\mathbf{x}_i,\mathbf{x})}+b,
\end{aligned}
\end{equation*}
where $\kappa$ denotes the kernel function. By using the kernel matrix as the bridge, the regularizers designed for the mapping function can be incorporated into the classifier's objective function. In this way, the final optimization problem is usually about the parameter (e.g., $\alpha_i$) or the kernel function. For example, the paper by Long {\it et al.} proposes a general framework termed Adaptation Regularization Based Transfer Learning (ARTL) \cite{LWD2014TKDE}. The goals of ARTL are to learn the adaptive classifier, to minimize the structural risk, to jointly reduce the marginal and the conditional distribution difference, and to maximize the manifold consistency between the data structure and the predictive structure. The authors also proposed two specific algorithms under this framework based on different loss functions. In these two algorithms, the coefficient matrix for computing MMD and the graph Laplacian matrix for manifold regularization are constructed at first. Then, a kernel function is selected to construct the kernel matrix. After that, the classifier learning problem is converted into a parameter (i.e., $\alpha_i$) solving problem, and the solution formula is also given in \cite{LWD2014TKDE}.

In ARTL, the choice of the kernel function affects the performance of the final classifier. In order to construct a robust classifier, some studies turn to kernel learning. For example, the paper by Duan {\it et al.} proposes a unified framework termed Domain Transfer Multiple Kernel Learning (DTMKL) \cite{DTX2012TPAMI}. In DTMKL, the kernel function is assumed to be a linear combination of a group of base kernels, i.e., $\kappa(\mathbf{x}_{i},\mathbf{x}_{j})=\sum_{k=1}^{N}\beta_{k}\kappa_{k}(\mathbf{x}_{i},\mathbf{x}_{j})$. DTMKL aims to minimize the distribution difference, the classification error, etc., simultaneously. The general objective function of DTMKL can be written as follows:
\begin{equation*}
\mathop{\min}_{\beta_{k},f}{\sigma\big(\text{MMD}(X^S,X^T;\kappa)\big)+\lambda\Omega^{L}(\beta_{k},f)},
\end{equation*}
where $\sigma$ is any monotonically increasing function, $f$ is the decision function with the same definition as the one in ARTL, and $\Omega^{L}(\beta_{k},f)$ is a general term representing a group of regularizers defined on the labeled instances such as the ones for minimizing the classification error and controlling the complexity of the resultant model. The authors developed an algorithm to learn the kernel and the decision function simultaneously by using the reduced gradient descent method \cite{RBC2008JMLR}. In each iteration, the weight coefficients of base kernels are fixed to update the decision function at first. Then, the decision function is fixed to update the weight coefficients. Note that DTMKL can incorporate many existing kernel methods. The authors proposed two specific algorithms under this framework. The first one implements the framework by using hinge loss and Support Vector Machine (SVM). The second one is an extension of the first one with an additional regularizer utilizing pseudo-label information, and the pseudo labels of the unlabeled instances are generated by using base classifiers.

\subsubsection{Feature Clustering} \label{SEC.FE.FC}
Feature clustering aims to find a more abstract feature representation of the original features. Although it can be regarded as a way of feature extraction, it is different from the above-mentioned mapping-based extraction.

For example, some transfer learning approaches implicitly reduce the features by using the co-clustering technique, i.e., simultaneously clustering both the columns and rows of (or say, co-cluster) a contingency table based on the information theory \cite{DMM2003KDD}. The paper by Dai {\it et al.} \cite{DXY2007KDD} proposes an algorithm termed Co-Clustering Based Classification (CoCC), which is used for document classification. In a document classification problem, the transfer learning task is to classify the target-domain documents (represented by a document-to-word matrix) with the help of the labeled source document-to-word data. CoCC regards the co-clustering technique as a bridge to transfer the knowledge. In CoCC algorithm, both the source and the target document-to-word matrices are co-clustered. The source document-to-word matrix is co-clustered to generate the word clusters based on the known label information, and these word clusters are used as constraints during the co-clustering process of the target-domain data. The co-clustering criterion is to minimize the loss in mutual information, and the clustering results are obtained by iteration. Each iteration contains the following two steps.
\begin{itemize}
\item[1.] {\it Document Clustering}: Each row of the target document-to-word matrix is re-ordered based on the objective function for updating the document clusters.

\item[2.] {\it Word Clustering}: The word clusters are adjusted to minimize the joint mutual-information loss of the source and the target document-word matrices.
\end{itemize}
After several times of iterations, the algorithm converges, and the classification results are obtained. Note that, in CoCC, the word clustering process implicitly extracts the word features to form unified word clusters.

Dai {\it et al.} also proposed an unsupervised clustering approach, which is termed as Self-Taught Clustering (STC) \cite{DYX2008ICML}. Similar to CoCC, this algorithm is also a co-clustering-based one. However, STC does not need the label information. STC aims to simultaneously co-cluster the source-domain and the target-domain instances with the assumption that these two domains share the same feature clusters in their common feature space. Therefore, two co-clustering tasks are separately performed at the same time to find the shared feature clusters. Each iteration of STC has the following steps.
\begin{itemize}
\item[1.] {\it Instance Clustering}: The clustering results of the source-domain and the target domain instances are updated to minimize their respective loss in mutual information.

\item[2.] {\it Feature Clustering}: The feature clusters are updated to minimize the joint loss in mutual information.
\end{itemize}
When the algorithm converges, the clustering results of the target-domain instances are obtained.

Different from the above-mentioned co-clustering-based ones, some approaches extract the original features into concepts (or topics). In the document classification problem, the concepts represent the high-level abstractness of the words (e.g., word clusters). In order to introduce the concept-based transfer learning approaches easily, let us briefly review the Latent Semantic Analysis (LSA) \cite{DDF1990JASIS}, the Probabilistic LSA (PLSA) \cite{H1999UAI}, and the Dual-PLSA \cite{YC2009ICASSP}.
\begin{itemize}[leftmargin=*]
\item {\it LSA}: LSA is an approach to mapping the document-to-word matrix to a low-dimensional space (i.e., a latent semantic space) based on the SVD technique. In short, LSA attempts to find the true meanings of the words. To realize this, SVD technique is used to reduce the dimensionality, which can remove the irrelevant information and filter out the noise information from the raw data.

\item {\it PLSA}: PLSA is developed based on a statistical view of LSA. PLSA assumes that there is a latent class variable $z$, which reflects the concept, associating the document $d$ and the word $w$. Besides, $d$ and $w$ are independently conditioned on the concept $z$. The diagram of this graphical model is presented as follows:
\begin{equation*}
	d\xleftarrow{P(d_{i}|z_{k})}{\overset{\underset{\Downarrow}{P(z_{k})}}z}\xrightarrow{P(w_{j}|z_{k})}w,
\end{equation*}
where the subscripts $i,j$ and $k$ represent the indexes of the document, the word, and the concept, respectively. PLSA constructs a Bayesian network, and the parameters are estimated by using the Expectation-Maximization (EM) algorithm \cite{DLR1977JRSC}.

\item {\it Dual-PLSA}: The Dual-PLSA is an extension of PLSA. This approach assumes there are two latent variables $z^{d}$ and $z^{w}$ associating the documents and the words. Specifically, the variables $z^{d}$ and $z^{w}$ reflect the concepts behind the documents and the words, respectively. The diagram of the Dual-PLSA is provided below:
\begin{equation*}
	d\xleftarrow{P(d_{i}|z^{d}_{k_1})}{z^{d}}\xleftrightarrow{P(z^{d}_{k_1},z^{w}_{k_2})}z^{w}\xrightarrow{P(w_{j}|z^{w}_{k_2})}w.
\end{equation*}
The parameters of the Dual-PLSA can also be obtained based on the EM algorithm.

\end{itemize}

Some concept-based transfer learning approaches are established based on PLSA. For example, the paper by Xue {\it et al.} proposes a cross-domain text classification approach termed Topic-Bridged Probabilistic Latent Semantic Analysis (TPLSA) \cite{XDY2008SIGIR}. TPLSA, which is an extension of PLSA, assumes that the source-domain and the target-domain instances share the same mixing concepts of the words. Instead of performing two PLSAs for the source domain and the target domain separately, the authors merge those two PLSAs as an integrated one by using the mixing concept $z$ as a bridge, i.e., each concept has some probabilities to produce the source-domain and the target-domain documents. The diagram of TPLSA is provided below:
\begin{equation*}
{\textcolor{white}{\Big\vert}}^{{d^{S}}}_{{d^{T}}}{}^\nwarrow_{\swarrow}\hspace{-0.42em}\xlongequal[P(d^{T}_{i}|z_{k})]{P(d^{S}_{i}|z_{k})}z\xleftarrow{P(z_{k}|w_j)}w.
\end{equation*}
Note that PLSA does not require the label information. In order to exploit the label information, the authors add the concept constraints, which include must-link and cannot-link constraints, as the penalty terms in the objective function of TPLSA. Finally, the objective function is iteratively optimized to obtain the classification results (i.e., ${\arg\max}_{z}{P(z|d_{i}^{T})}$) by using the EM algorithm.

The work by Zhuang {\it et al.} proposes an approach termed Collaborative Dual-PLSA (CD-PLSA) for multi-domain text classification ($m^{S}$ source domains and $m^{T}$ target domains) \cite{ZLS2010CIKM,ZLS2012TKDE}. CD-PLSA is an extension of Dual-PLSA. Its diagram is shown below:
\begin{equation*}
\myunderrightnarrow{{\overset{\underset{\Downarrow}{P(\mathcal{D}_{k_0})}}{\mathcal{D}}}\rightarrow{\overset{\underset{\Downarrow}{P(d_{i}|z^{d}_{k_1},\mathcal{D}_{k_0})}}d}{\leftarrow{z^{d}}}{\xleftrightarrow{P(z^{d}_{k_1},z^{w}_{k_2})}}z^{w}\rightarrow{\overset{\underset{\Downarrow}{P(w_{j}|z^{w}_{k_2},\mathcal{D}_{k_0})}}w}},
\end{equation*}
where $1\leq k_0\leq m^{S}+m^{T}$ denotes the domain index. The domain $\mathcal{D}$ connects both the variables $d$ and $w$, but is independent of the variables $z^{d}$ and $z^{w}$. The label information of the source-domain instances is utilized by initializing the value $P(d_{i}|z^{d}_{k_1},\mathcal{D}_{k_0})$ ($k_0=1,\cdots,m^{S}$). Due to the lack of the target-domain label information, the value $P(d_{i}|z^{d}_{k_1},\mathcal{D}_{k_0})$ ($k_0=m^{S}+1,\cdots,m^{S}+m^{T}$) can be initialized based on any supervised classifier. Similarly, the authors adopt the EM algorithm to find the parameters. Through the iterations, all the parameters in the Bayesian network are obtained. Thus, the class label of the $i$-th document in a target domain (denoted by $\mathcal{D}_{k}$) can be predicted by computing the posterior probabilities, i.e., ${\arg\max}_{z^{d}}{P(z^{d}|d_{i},\mathcal{D}_{k})}$.

Zhuang {\it et al.} further proposed a general framework that is termed as Homogeneous-Identical-Distinct-Concept Model (HIDC) \cite{ZLY2013IJCAI}. This framework is also an extension of Dual-PLSA. HIDC is composed of three generative models, i.e., identical-concept, homogeneous-concept, and distinct-concept models. These three graphical models are presented below:
\begin{align*}
&\text{Identical-Concept Model:}&\myunderright{\mathcal{D}\rightarrow d\leftarrow{z^{d}}}\rightarrow z_{\text{IC}}^{w}\rightarrow w,\\
&\text{Homogeneous-Concept Model:}&\myoverright{\myunderright{\mathcal{D}\rightarrow d\leftarrow{z^{d}}}\rightarrow z_{\text{HC}}^{w}\rightarrow w},\\
&\text{Distinct-Concept Model:}&\mydoubleright{\myunderright{\mathcal{D}\rightarrow d\leftarrow{z^{d}}}\rightarrow z_{\text{DC}}^{w}\rightarrow w}.
\end{align*}
The original word concept $z^{w}$ is divided into three types, i.e., $z_{\text{IC}}^{w}$, $z_{\text{HC}}^{w}$, and $z_{\text{DC}}^{w}$. In the identical-concept model, the word distributions only rely on the word concepts, and the word concepts are independent of the domains. However, in the homogeneous-concept model, the word distributions also depend on the domains. The difference between the identical and the homogeneous concepts is that $z_{\text{IC}}^{w}$ is directly transferable, while $z_{\text{HC}}^{w}$ is the domain-specific transferable one that may have different effects on the word distributions for different domains. In the distinct-concept model, $z_{\text{DC}}^{w}$ is actually the nontransferable domain-specific one, which may only appear in a specific domain. The above-mentioned three models are combined as an integrated one, i.e., HIDC. Similar to other PLSA-related algorithms, HIDC also uses EM algorithm to get the parameters.

\subsubsection{Feature Selection} \label{SEC.FE.SL}
Feature selection is another kind of operation for feature reduction, which is used to extract the pivot features. The pivot features are the ones that behave in the same way in different domains. Due to the stability of these features, they can be used as the bridge to transfer the knowledge. For example, Blitzer {\it et al.} proposed an approach termed Structural Correspondence Learning (SCL) \cite{BMP2006EMNLP}. Briefly, SCL consists of the following steps to construct a new feature representation.
\begin{itemize}
\item[1.] {\it Feature Selection}: SCL first performs feature selection operations to obtain the pivot features.
\item[2.] {\it Mapping Learning}: The pivot features are utilized to find a low-dimensional common latent feature space by using the structural learning technique \cite{AZ2005JMLR}.
\item[3.] {\it Feature Stacking}: A new feature representation is constructed by feature augmentation, i.e., stacking the original features with the obtained low-dimensional features.
\end{itemize}
Take the part-of-speech tagging problem as an example. The selected pivot features should occur frequently in source and target domains. Therefore, determiners can be included in pivot features. Once all the pivot features are defined and selected, a number of binary linear classifiers whose function is to predict the occurrence of each pivot feature are constructed. Without losing generality, the decision function of the $i$-th classifier, which is used to predict the $i$-th pivot feature, can be formulated as $f_{i}(\mathbf{x})=\text{sign}({\bm\theta}_i\cdot\mathbf{x})$, where $\mathbf{x}$ is assumed to be a binary feature input. And the $i$-th classifier is trained on all the instances excluding the features derived from the $i$-th pivot feature. The following formula can be used to estimate the $i$-th classifier's parameters, i.e.,
\begin{equation*}
{\bm\theta}_i=\mathop{\arg\min}_{\bm\theta}{\dfrac{1}{n}\sum_{j=1}^{n}{\mathcal{L}\left({\bm\theta}\cdot\mathbf{x}_j,\text{Row}_{i}({\mathbf{x}_j})\right)}+\lambda||\bm\theta||^2},
\end{equation*}
where $\text{Row}_{i}({\mathbf{x}_j})$ denotes the true value of the unlabeled instance $\mathbf{x}_j$ in terms of the $i$-th pivot feature. By stacking the obtained parameter vectors as column elements, a matrix $\tilde{W}$ is obtained. Next, based on singular value decomposition (SVD), the top-$k$ left singular vectors, which are the principal components of the matrix $\tilde{W}$, are taken to construct the transformation matrix $W$. At last, the final classifier is trained on the labeled instances in an augmented feature space, i.e., $([\mathbf{x}^{L}_i;W^{\text{T}}\mathbf{x}^{L}_i]^{\text{T}},y^{L}_i).$

\subsubsection{Feature Encoding} \label{SEC.FE.EC}
In addition to feature extraction and selection, feature encoding is also an effective tool. For example, autoencoders, which are often adopted in deep learning area, can be used for feature encoding. An autoencoder consists of an encoder and a decoder. The encoder tries to produce a more abstract representation of the input, while the decoder aims to map back that representation and to minimize the reconstruction error. Autoencoders can be stacked to build a deep learning architecture. Once an autoencoder completes the training process, another autoencoder can be stacked at the top of it. The newly added autoencoder is then trained by using the encoded output of the upper-level autoencoder as its input. In this way, deep learning architectures can thus be constructed.

Some transfer learning approaches are developed based on autoencoders. For example, the paper by Glorot {\it et al.} proposes an approach termed Stacked Denoising Autoencoder (SDA) \cite{GBB2011ICML}. The denoising autoencoder, which can enhance the robustness, is an extension of the basic one \cite{VLB2008ICML}. This kind of autoencoder contains a randomly corrupting mechanism that adds noise to the input before mapping. For example, an input can be corrupted or partially destroyed by adding a masking noise or Gaussian noise. The denoising autoencoder is then trained to minimize the denoising reconstruction error between the original clean input and the output. The SDA algorithm proposed in the paper mainly encompasses the following steps.
\begin{itemize}
\item[1.] {\it Autoencoder Training}: The source-domain and target-domain instances are used to train a stack of denoising autoencoders in a greedy layer-by-layer way.
\item[2.] {\it Feature Encoding $\&$ Stacking}: A new feature representation is constructed by stacking the encoding output of intermediate layers, and the features of the instances are transformed into the obtained new representation.
\item[3.] {\it Learner Training}: The target classifier is trained on the transformed labeled instances.
\end{itemize}

Although the SDA algorithm has excellent performance for feature extraction, it still has some drawbacks such as high computational and parameter-estimation cost. In order to shorten the training time and to speed up traditional SDA algorithms, Chen {\it et al.} proposed a modified version of SDA, i.e., Marginalized Stacked Linear Denoising Autoencoder (mSLDA) \cite{CXW2012ICML,CWX2015JMLR}. This algorithm adopts linear autoencoders and marginalizes the randomly corrupting step in a closed form. It may seem that linear autoencoders are too simple to learn complex features. However, the authors observe that linear autoencoders are often sufficient to achieve competent performance when encountering high dimensional data. The basic architecture of mSLDA is a single-layer linear autoencoder. The corresponding single-layer mapping matrix $W$ (augmented with a bias column for convenience) should minimize the expected squared reconstruction loss function, i.e.,
\begin{equation*}
W = \mathop{\arg\min}_{W	}{\dfrac{1}{2n}\sum_{i=1}^{n}\mathbb{E}_{P(\tilde{\mathbf{x}}_i|\mathbf{x})}\left[\vert\vert\mathbf{x}_i-W\tilde{\mathbf{x}}_i\vert\vert^{2}\right]},
\end{equation*}
where $\tilde{\mathbf{x}}_i$ denotes the corrupted version of the input $\mathbf{x}_i$. The solution of $W$ is given by \cite{CXW2012ICML,CWX2015JMLR}:
\begin{equation*}
W=\left(\sum_{i=1}^{n}{\mathbf{x}_{i}\mathbb{E}[\tilde{\mathbf{x}}_i]^{\rm T}}\right)\left(\sum_{i=1}^{n}{\mathbb{E}\left[\tilde{\mathbf{x}}_i\tilde{\mathbf{x}}_{i}^{\rm T}\right]}\right)^{-1}.
\end{equation*}
When the corruption strategy is determined, the above formulas can be further expanded and simplified into a specific form. Note that, in order to insert nonlinearity, a nonlinear function is used to squash the output of each autoencoder after we obtain the matrix $W$ in a closed form. Then, the next linear autoencoder is stacked to the current one in a similar way to SDA. In order to deal with high dimensional data, the authors also put forward an extension approach to further reduce the computational complexity.

\subsubsection{Feature Alignment}
Note that feature augmentation and feature reduction mainly focus on the explicit features in a feature space. In contrast, in addition to the explicit features, feature alignment also focuses on some implicit features such as the statistic features and the spectral features. Therefore, feature alignment can play various roles in the feature transformation process. For example, the explicit features can be aligned to generate a new feature representation, or the implicit features can be aligned to construct a satisfied feature transformation.

There are several kinds of features that can be aligned, which includes subspace features, spectral features, and statistic features. Take the subspace feature alignment as an example. A typical approach mainly has the following steps.
\begin{itemize}
\item[1.] {\it Subspace Generation}: In this step, the instances are used to generate the respective subspaces for the source and the target domains. The orthonormal bases of the source and the target domain subspaces are then obtained, which are denoted by $M_{S}$ and $M_{T}$, respectively. These bases are used to learn the shift between the subspaces.
\item[2.] {\it Subspace Alignment}: In the second step, a mapping, which aligns the bases $M_{S}$ and $M_{T}$ of the subspaces, is learned. And the features of the instances are projected to the aligned subspaces to generate new feature representation.
\item[3.] {\it Learner Training}: Finally, the target learner is trained on the transformed instances.
\end{itemize}
For example, the work by Fernando {\it et al.} proposes an approach termed Subspace Alignment (SA) \cite{FHS2013ICCV}. In SA, the subspaces are generated by performing PCA; the bases $M_{S}$ and $M_{T}$ are obtained by selecting the leading eigenvectors. Then, a transformation matrix $W$ is learned to align the subspaces, which is given by \cite{FHS2013ICCV}:
\begin{equation*}
W=\mathop{\arg\min}_{W}{\vert\vert M_{S}W-M_{T}\vert\vert^{2}_{F}}=M_{S}^{\text{T}}M_{T},
\end{equation*}
where $\vert\vert\cdot\vert\vert_{F}$ denotes the Frobenius norm. Note that the matrix $W$ aligns $M_{S}$ with $M_{T}$, or say, transforms the source subspace coordinate system into the target subspace coordinate system. The transformed low-dimensional source-domain and target-domain instances are given by $X^{S}M_{S}W$ and $X^{T}M_{T}$, respectively. Finally, a learner can be trained on the resultant transformed instances.

In light of SA, a number of transfer learning approaches are established. For example, the paper by Sun and Saenko proposes an approach that aligns both the subspace bases and the distributions \cite{SS2015BMVC}, which is termed as Subspace Distribution Alignment between Two Subspaces (SDA-TS). In SDA-TS, the transformation matrix $W$ is formulated as $W=M_{S}^{\text{T}}M_{T}Q$, where $Q$ is a matrix used to align the distribution difference. The transformation matrix $W$ in SA is a special case of the one in SDA-TS by setting $Q$ to an identity matrix. Note that SA is a symmetrical feature-based approach, while SDA-TS is an asymmetrical one. In SDA-TS, the labeled source-domain instances are projected to the source subspace, then mapped to the target subspace, and finally mapped back to the target domain. The transformed source-domain instances are formulated as $X^{S}M_{S}WM_{T}^{\text{T}}$.

Another representative subspace feature alignment approach is Geodesic Flow Kernel (GFK) \cite{G2012CVPR}, which is proposed by Gong {\it et al}. GFK is closely related to a previous approach termed Geodesic Flow Subspaces (GFS) \cite{GLC2011CVPR}. Before introducing GFK, let us review the steps of GFS at first. GFS is inspired by incremental learning. Intuitively, utilizing the information conveyed by the potential path between two domains may be beneficial to the domain adaptation. GFS generally takes the following steps to align features.
\begin{itemize}
\item[1.] {\it Subspace Generation}: GFS first generates two subspaces of the source and the target domains by performing PCA, respectively.
\item[2.] {\it Subspace Interpolation}: The two obtained subspaces can be viewed as two points on the Grassmann manifold \cite{Z2000EMS}. A finite number of the interpolated subspaces are generated between these two subspaces based on the geometric properties of the manifold.
\item[3.] {\it Feature Projection $\&$ Stacking}: The original features are transformed by stacking the corresponding projections from all the obtained subspaces.
\end{itemize}
Despite the usefulness and superiority of GFS, there is a problem about how to determine the number of the interpolated subspaces. GFK resolves this problem by integrating infinite number of the subspaces located on the geodesic curve from the source subspace to the target one. The key of GFK is to construct an infinite-dimensional feature space that incorporating the information of all the subspaces lying on the geodesic flow. In order to compute the inner product in the resultant infinite-dimensional space, the geodesic-flow kernel is defined and derived. In addition, a subspace-disagreement measure is proposed to select the optimal dimensionality of the subspaces; a rank-of-domain metric is also proposed to select the optimal source domain when multi-source domains are available.

Statistic feature alignment is another kind of feature alignment. For example, Sun {\it et al.} proposed an approach termed Co-Relation Alignment (CORAL) \cite{SFS2016AAAI}. CORAL constructs the transformation matrix of the source features by aligning the second-order statistic features, i.e., the covariance matrices. The transformation matrix $W$ is given by \cite{SFS2016AAAI}:
\begin{equation*}
W=\mathop{\arg\min}_{W}{\vert\vert W^{\text{T}}C_{S}W-C_{T}\vert\vert^{2}_{F}},
\end{equation*}
where $C$ denotes the covariance matrix. Note that, compared to the above subspace-based approaches, CORAL avoids subspace generation as well as projection and is very easy to implement.

Some transfer learning approaches are established based on spectral feature alignment. In traditional machine learning area, spectral clustering is a clustering technique based on graph theory. The key of this technique is to utilize the spectrum, i.e., eigenvalues, of the similarity matrix to reduce the dimension of the features before clustering. The similarity matrix is constructed to quantitatively assess the relative similarity of each pair of data/vertices. On the basis of spectral clustering and feature alignment, Spectral Feature Alignment (SFA) \cite{PNS2010WWW} is proposed by Pan {\it et al}. SFA is an algorithm for sentiment classification. This algorithm tries to identify the domain-specific words and domain-independent words in different domains, and then aligns these domain-specific word features to construct a low-dimensional feature representation. SFA generally contains the following five steps.
\begin{itemize}
\item[1.] {\it Feature Selection}: In this step, feature selection operations are performed to select the domain-independent/pivot features. The paper presents three strategies to select domain-independent features. These strategies are based on the occurrence frequency of words, the mutual information between features and labels \cite{BDP2007ACL}, and the mutual information between features and domains, respectively.
\item[2.] {\it Similarity Matrix Construction}: Once the domain-specific and the domain-independent features are identified, a bipartite graph is constructed. Each edge of this bipartite graph is assigned with a weight that measures the co-occurrence relationship between a domain-specific word and a domain-independent word. Based on the bipartite graph, a similarity matrix is then constructed.
\item[3.] {\it Spectral Feature Alignment}: In this step, a spectral clustering algorithm is adapted and performed to align domain-specific features \cite{C1997AMS,NJW2001NIPS}. Specifically, based on the eigenvectors of the graph Laplacian, a feature alignment mapping is constructed, and the domain-specific features are mapped into a low-dimensional feature space.
\item[4.] {\it Feature Stacking}: The original features and the low-dimensional features are stacked to produce the final feature representation.
\item[5.] {\it Learner Training}: The target learner is trained on the labeled instances with the final feature representation.
\end{itemize}

There are some other spectral transfer learning approaches. For example, the work by Ling {\it et al.} proposes an approach termed Cross-Domain Spectral Classifier (CDSC) \cite{LDX2008KDD}. The general ideas and steps of this approach are presented as follows.
\begin{itemize}
\item[1.] {\it Similarity Matrix Construction}: In the first step, two similarity matrices are constructed corresponding to the whole instances and the target-domain instances, respectively.

\item[2.] {\it Spectral Feature Alignment}: An objective function is designed with respect to a graph-partition indicator vector; a constraint matrix is constructed, which contains pair-wise must-link information. Instead of seeking the discrete solution of the indicator vector, the solution is relaxed to be continuous, and the eigen-system problem corresponding to the objective function is solved to construct the aligned spectral features \cite{KKM2003IJCAI}.

\item[3.] {\it Learner Training}: A traditional classifier is trained on the transformed instances.
\end{itemize}
To be more specific, the objective function has a form of the generalized Rayleigh quotient, which aims to find the optimal graph partition that respects the label information with small cut-size \cite{SM2000TPAMI}, to maximize the separation of the target-domain instances, and to fit the constraints of the pair-wise property. After eigen-decomposition, the last eigenvectors are selected and combined as a matrix, and then the matrix is normalized. Each row of the normalized matrix represents a transformed instance.

\section{Model-Based Interpretation} \label{SEC.MBI}
\begin{figure*}
\centering
\includegraphics[width=1.5\columnwidth]{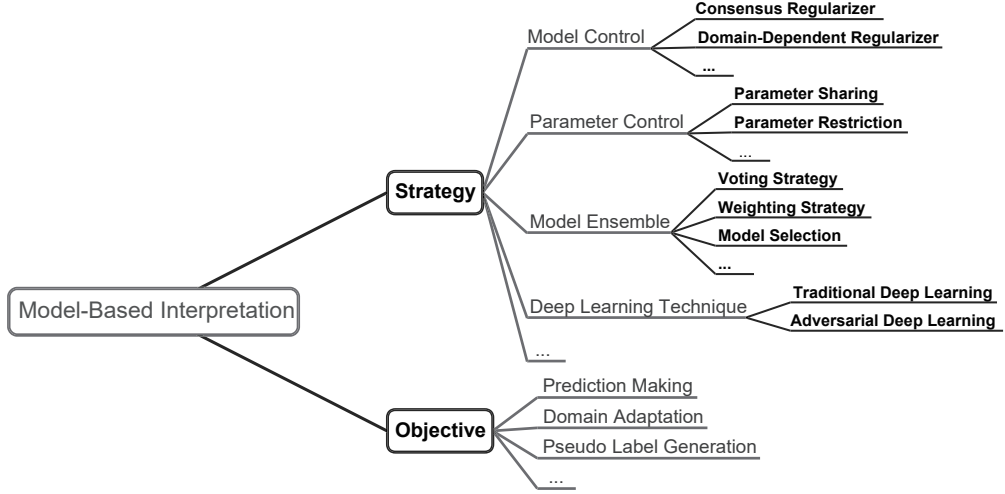}
\caption{Strategies and objectives of the transfer learning approaches from the model perspective.}
\label{Fig3}
\end{figure*}
Transfer learning approaches can also be interpreted from the model perspective. Fig. \ref{Fig3} shows the corresponding strategies and the objectives. The main objective of a transfer learning model is to make accurate prediction results on the target domain, e.g., classification or clustering results. Note that a transfer learning model may consist of a few sub-modules such as classifiers, extractors, or encoders. These sub-modules may play different roles, e.g., feature adaptation or pseudo label generation. In this section, some related transfer learning approaches are introduced in proper order according to the strategies shown in Fig. \ref{Fig3}.

\subsection{Model Control Strategy}
From the perspective of model, a natural thought is to directly add the model-level regularizers to the learner's objective function. In this way, the knowledge contained in the pre-obtained source models can be transferred into the target model during the training process. For example, the paper by Duan {\it et al.} proposes a general framework termed Domain Adaptation Machine (DAM) \cite{DTX2009ICML,DXT2012TNNLS}, which is designed for multi-source transfer learning. The goal of DAM is to construct a robust classifier for the target domain with the help of some pre-obtained base classifiers that are respectively trained on multiple source domains. The objective function is given by:
\begin{equation*}
\min_{f^{T}}{{\mathcal{L}}^{T,L}(f^{T})+\lambda_{1}\Omega^{\text{D}}(f^{T})+\lambda_2\Omega(f^{T})},
\end{equation*}
where the first term represents the loss function used to minimize the classification error of the labeled target-domain instances, the second term denotes different regularizers, and the third term is used to control the complexity of the final decision function $f^{T}$. Different types of the loss functions can be adopted in ${\mathcal{L}}^{T,L}(f^{T})$ such as the square error or the cross-entropy loss. Some transfer learning approaches can be regarded as the special cases of this framework to some extent.
\begin{itemize}[leftmargin=*]
\item (Consensus Regularizer) The work by Luo {\it et al.} proposes a framework termed Consensus Regularization Framework (CRF) \cite{LZX2008CIKM,ZLX2010TKDE}. CRF is designed for multi-source transfer learning with no labeled target-domain instances. The framework constructs $m^{S}$ classifiers corresponding to each source domain, and these classifiers are required to reach mutual consensuses on the target domain. The objective function of each source classifier, denoted by $f_k^{S}$ (with $k=1,\cdots,m^{S}$), is similar to that of DAM, which is presented below:
\begin{equation*}
	\begin{aligned}
		\min_{f^{S}_{k}}&{-\sum_{i=1}^{n^{S_{k}}}\log P(y_{i}^{S_{k}}|\mathbf{x}_{i}^{S_{k}};f^{S}_{k})+\lambda_{2}\Omega(f_{k}^{S})}\\&{+\lambda_{1}\sum_{i=1}^{n^{T,U}}\sum_{y_{j}\in\mathcal{Y}}{S}\Big(\dfrac{1}{m^{S}}\sum_{k_{0}=1}^{m^{S}}P(y_{j}|\mathbf{x}^{T,U}_{i};f_{k_{0}}^{S})\Big)},
	\end{aligned}
\end{equation*}
where $f^{S}_k$ denotes the decision function corresponding to the $k$-th source domain, and ${S}(x)=-x\log x$. The first term is used to quantify the classification error of the $k$-th classifier on the $k$-th source domain, and the last term is the consensus regularizer in the form of cross-entropy. The consensus regularizer can not only enhance the agreement of all the classifiers, but also reduce the uncertainty of the predictions on the target domain. The authors implement this framework based on the logistic regression. A difference between DAM and CRF is that DAM explicitly constructs the target classifier, while CRF makes the target predictions based on the reached consensus from the source classifiers.

\item (Domain-dependent Regularizer) Fast-DAM is a specific algorithm of DAM \cite{DTX2009ICML}. In light of the manifold assumption \cite{BNS2006JMLR} and the graph-based regularizer \cite{EMP2005JMLR,KKS2007NIPS}, Fast-DAM designs a domain-dependent regularizer. The objective function is given by:
\begin{equation*}
	\begin{aligned}
		\min_{f^{T}}&~{\sum_{j=1}^{n^{T,L}}\left(f^{T}(\mathbf{x}^{T,L}_j)-y^{T,L}_j\right)^2+\lambda_{2}\Omega(f^{T})}\\&{+\lambda_{1}\sum_{k=1}^{m^{S}}\beta_{k}\sum_{i=1}^{n^{T,U}}{\left(f^{T}(\mathbf{x}^{T,U}_i)-f^{S}_{k}(\mathbf{x}^{T,U}_i)\right)^2}},
	\end{aligned}
\end{equation*}
where $f^{S}_k$ ($k=1,2,\cdots,m^{S}$) denotes the pre-obtained source decision function for the $k$-th source domain and $\beta_{k}$ represents the weighting parameter that is determined by the relevance between the target domain and the $k$-th source domain and can be measured based on the MMD metric. The third term is the domain-dependent regularizer, which transfers the knowledge contained in the source classifier motivated by domain dependence. In \cite{DTX2009ICML}, the authors also introduce and add a new term to the above objective function based on $\varepsilon$-insensitive loss function \cite{SS2004SC}, which makes the resultant model have high computational efficiency.

\item (Domain-dependent Regularizer $+$ Universum Regularizer) Univer-DAM is an extension of the Fast-DAM \cite{DXT2012TNNLS}. Its objective function contains an additional regularizer, i.e., Universum regularizer. This regularizer usually utilizes an additional dataset termed Universum where the instances do not belong to either the positive or the negative class \cite{WCS2006ICML}. The authors treat the source-domain instances as the Universum for the target domain, and the objective function of Univer-DAM is presented as follows:
\begin{align*}
	\min_{f^{T}}&~{\sum_{j=1}^{n^{T,L}}\left(f^{T}(\mathbf{x}^{T,L}_j)-y^{T,L}_j\right)^2+{\lambda_{2}}{\sum_{j=1}^{n^{S}}\left(f^{T}(\mathbf{x}^{S}_j)\right)^2}}\\&{+{\lambda_1}\sum_{k=1}^{m^{S}}\beta_{k}\sum_{i=1}^{n^{T,U}}{\left(f^{T}(\mathbf{x}^{T,U}_i)-f^{S}_{k}(\mathbf{x}^{T,U}_i)\right)^2}+\lambda_{3}\Omega(f^{T})}.
\end{align*}
Similar to Fast-DAM, the $\varepsilon$-insensitive loss function can also be utilized \cite{DXT2012TNNLS}.

\end{itemize}

\subsection{Parameter Control Strategy}
The parameter control strategy focuses on the parameters of models. For example, in the application of object categorization, the knowledge from known source categories can be transferred into target categories via object attributes such as shape and color \cite{YA2010ECCV}. The attribute priors, i.e., probabilistic distribution parameters of the image features corresponding to each attribute, can be learned from the source domain and then used to facilitate learning the target classifier. The parameters of a model actually reflect the knowledge learned by the model. Therefore, it is possible to transfer the knowledge at the parametric level.

\subsubsection{Parameter Sharing}
An intuitive way of controlling the parameters is to directly share the parameters of the source learner to the target learner. Parameter sharing is widely employed especially in the network-based approaches. For example, if we have a neural network for the source task, we can freeze (or say, share) most of its layers and only finetune the last few layers to produce a target network. The network-based approaches are introduced in Section \ref{SEC.DL}.

In addition to network-based parameter sharing, matrix-factorization-based parameter sharing is also workable. For example, Zhuang {\it et al.} proposed an approach for text classification, which is referred to as Matrix Tri-Factorization Based Classification Framework (MTrick) \cite{ZLX2011SADM}. The authors observe that, in different domains, different words or phrases sometimes express the same or similar connotative meaning.
Thus, it is more effective to use the concepts behind the words rather than the words themselves as a bridge to transfer the knowledge in source domains. Different from PLSA-based transfer learning approaches that utilize the concepts by constructing Bayesian networks, MTrick attempts to find the connections between the document classes and the concepts conveyed by the word clusters through matrix tri-factorization. These connections are considered to be the stable knowledge that is supposed to be transferred. The main idea is to decompose a document-to-word matrix into three matrices, i.e., document-to-cluster, connection, and cluster-to-word matrices. Specifically, by performing the matrix tri-factorization operations on the source and the target document-to-word matrices respectively, a joint optimization problem is constructed, which is given by
\begin{equation*}
\begin{aligned}
	&\min_{Q,R,W}{||X^{S}-Q^{S}R{W^{S}}||^{2}+\lambda_{1}||X^{T}-Q^{T}RW^{T}||^{2}}\\&\qquad\ ~{+\lambda_{2}||Q^{S}-\breve{Q}^{S}||^2}\\
	&{~~s.t.~}\text{Normalization Constraints},
\end{aligned}
\end{equation*}
where $X$ denotes the document-to-word matrix, $Q$ denotes the document-to-cluster matrix, $R$ represents the transformation matrix from document clusters to word clusters, $W$ denotes the cluster-to-word matrix, $n^{d}$ denotes the number of the documents, and $\breve{Q}^{S}$ represents the label matrix. The matrix $\breve{Q}^{S}$ is constructed based on the class information of the source-domain documents. If the $i$-th document belongs to the $k$-th class, $\breve{Q}^{S}_{[i,k]}=1$. In the above objective function, the matrix $R$ is actually the shared parameter. The first term aims to tri-factorize the source document-to-word matrix, and the second term decomposes the target document-to-word matrix. The last term incorporates the source-domain label information. The optimization problem is solved based on the alternating iteration method. Once the solution of $Q^{T}$ is obtained, the class index of the $k$-th target-domain instance is the one with the maximum value in the $k$-th row of $Q^{T}$.

Further, Zhuang {\it et al.} extended MTrick and proposed an approach termed Triplex Transfer Learning (TriTL) \cite{ZLD2014TC}. MTrick assumes that the domains share the similar concepts behind their word clusters. In contrast, TriTL assumes that the concepts of these domains can be further divided into three types, i.e., domain-independent, transferable domain-specific, and nontransferable domain-specific concepts, which is similar to HIDC. This idea is motivated by Dual Transfer Learning (DTL), where the concepts are assumed to be composed of the domain-independent ones and the transferable domain-specific ones \cite{LWD2012ICDM}. The objective function of TriTL is provided as follows:
\begin{equation*}
\begin{aligned}
	&\min_{Q,R,W}{\sum_{k=1}^{m^{S}+m^{T}}||X_{k}-Q_{k}\begin{bmatrix}R^{\text{DI}}&R^{\text{TD}}&R^{\text{ND}}_{k}\end{bmatrix}\begin{bmatrix}W^{\text{DI}}\\W^{\text{TD}}_{k}\\W^{\text{ND}}_k\end{bmatrix}||^{2}}\\
	&~~s.t.~\text{Normalization Constraints},
\end{aligned}
\end{equation*}
where the definitions of the symbols are similar to those of MTrick and the subscript $k$ denotes the index of the domains with the assumption that the first $m^{S}$ domains are the source domains and the last $m^{T}$ domains are the target domains. The authors proposed an iterative algorithm to solve the optimization problem. And in the initialization phase, $W^{\text{DI}}$ and $W^{\text{TD}}_{k}$ are initialized based on the clustering results of the PLSA algorithm, while $W^{\text{UT}}_k$ is randomly initialized; the PLSA algorithm is performed on the combination of the instances from all the domains.

There are some other approaches developed based on matrix factorization. Wang {\it et al.} proposed a transfer learning framework for image classification \cite{WNH2011ICCV}. Wang {\it et al.} proposed a softly associative approach that integrates two matrix tri-factorizations into a joint framework \cite{WLW2019TC}. Do {\it et al.} utilized matrix tri-factorization to discover both the implicit and the explicit similarities for cross-domain recommendation \cite{DLF2019TKDE}.

\subsubsection{Parameter Restriction}
Another parameter-control-type strategy is to restrict the parameters. Different from the parameter sharing strategy that enforces the models share some parameters, parameter restriction strategy only requires the parameters of the source and the target models to be similar.

Take the approaches to category learning as examples. The category-learning problem is to learn a new decision function for predicting a new category (denoted by the $(k+1)$-th category) with only limited target-domain instances and $k$ pre-obtained binary decision functions. The function of these pre-obtained decision functions is to predict which of the $k$ categories an instance belongs to. In order to solve the category-learning problem, Tommasi {\it et al.} proposed an approach termed Single-Model Knowledge Transfer (SMKL) \cite{TC2009BMVC}. SMKL is based on Least-Squares SVM (LS-SVM). The advantage of LS-SVM is that LS-SVM transforms inequality constraints to equality constraints and has high computational efficiency; its optimization is equivalent to solving a linear equation system problem instead of a quadratic programming problem. SMKL selects one of the pre-obtained binary decision functions, and transfers the knowledge contained in its parameters. The objective function is given by
\begin{equation*}
\min_{f}{\dfrac{1}{2}\left\vert\left\vert\bm\theta-{\beta\tilde{\bm\theta}}\right\vert\right\vert^{2}+\dfrac{\lambda}{2}\sum_{j=1}^{n^{T,L}}\eta_{j}\left(f(\mathbf{x}^{T,L}_{j})-y^{T,L}_j\right)^2},
\end{equation*}
where $f(\mathbf{x})={\bm\theta}\cdot\Phi(\mathbf{x})+b$, $\beta$ is the weighting parameter controlling the transfer degree, $\tilde{\bm\theta}$ is the parameter of a selected pre-obtained model, and $\eta_{j}$ is the coefficient for resolving the label imbalance problem. The kernel parameter and the tradeoff parameter are chosen based on cross-validation. In order to find the optimal weighting parameter, the authors refer to an earlier work \cite{C2006IJCNN}. In \cite{C2006IJCNN}, Cawley proposed a model selection mechanism for LS-SVM, which is based on the leave-one-out cross-validation method. The superiority of this method is that the leave-one-out error for each instance can be obtained in a closed form without performing the real cross-validation experiment. Motivated by Cawley's work, the generalization error can be easily estimated to guide the parameter setting in SMKL.

Tommasi {\it et al.} further extended SMKL by utilizing all the pre-obtained decision functions. In \cite{TOC2010CVPR}, an approach that is referred to as Multi-Model Knowledge Transfer (MMKL) is proposed. Its objective function is presented as follows:
\begin{equation*}
\min_{f}{\dfrac{1}{2}\left\vert\left\vert\bm\theta-\sum_{i=1}^{k}{\beta_{i}\bm\theta_i}\right\vert\right\vert^{2}+\dfrac{\lambda}{2}\sum_{j=1}^{n^{T,L}}\eta_{j}\left(f(\mathbf{x}^{T,L}_{j})-y^{T,L}_j\right)^2},
\end{equation*}
where $\bm\theta_i$ and $\beta_{i}$ are the model parameter and the weighting parameter of the $i$-th pre-obtained decision function, respectively. The leave-one-out error can also be obtained in a closed form, and the optimal value of $\beta_{i}$ ($i=1,2,\cdots,k$) is the one that maximizes the generalization performance.

\subsection{Model Ensemble Strategy} \label{SEC.MES}
In sentiment analysis applications related to product reviews, data or models from multiple product domains are available and can be used as the source domains \cite{LLY2014CIKM}. Combining data or models directly into a single domain may not be successful because the distributions of these domains are different from each other. Model ensemble is another commonly used strategy. This strategy aims to combine a number of weak classifiers to make the final predictions. Some previously mentioned transfer learning approaches already adopted this strategy. For example, TrAdaBoost and MsTrAdaBoost ensemble the weak classifiers via voting and weighting, respectively. In this subsection, several typical ensemble-based transfer learning approaches are introduced to help readers better understand the function and the appliance of this strategy.

As mentioned in Section \ref{SEC.IWS}, TaskTrAdaBoost, which is an extension of TrAdaBoost for handling multi-source scenarios, is proposed in the paper \cite{YD2010CVPR}. TaskTrAdaBoost mainly has the following two stages.
\begin{itemize}
\item[1.] {\it Candidate Classifier Construction}: In the first stage, a group of candidate classifiers are constructed by performing AdaBoost on each source domain. Note that, for each source domain, each iteration of AdaBoost results in a new weak classifier. In order to avoid the over-fitting problem, the authors introduced a threshold to pick the suitable classifiers into the candidate group.
\item[2.] {\it Classifier Selection and Ensemble}: In the second stage, a revised version of AdaBoost is performed on the target-domain instances to construct the final classifier. In each iteration, an optimal candidate classifier which has the lowest classification error on the labeled target-domain instances is picked out and assigned with a weight based on the classification error. Then, the weight of each target-domain instance is updated based on the performance of the selected classifier on the target domain. After the iteration process, the selected classifiers are ensembled to produce the final predictions.
\end{itemize}
The difference between the original AdaBoost and the second stage of TaskTrAdaBoost is that, in each iteration, the former constructs a new candidate classifier on the weighted target-domain instances, while the latter selects one pre-obtained candidate classifier which has the minimal classification error on the weighted target-domain instances.

The paper by Gao {\it et al.} proposes another ensemble-based framework that is referred to as Locally Weighted Ensemble (LWE) \cite{GFJ2008KDD}. LWE focuses on the ensemble process of various learners; these learners could be constructed on different source domains, or be built by performing different learning algorithms on a single source domain. Different from TaskTrAdaBoost that learns the global weight of each learner, the authors adopted the local-weight strategy, i.e., assigning adaptive weights to the learners based on the local manifold structure of the target-domain test set. In LWE, a learner is usually assigned with different weights when classifying different target-domain instances. Specifically, the authors adopt a graph-based approach to estimate the weights. The steps for weighting are outlined below.
\begin{itemize}
\item[1.] {\it Graph Construction}: For the $i$-th source learner, a graph $G^{T}_{S_i}$ is constructed by using the learner to classify the target-domain instances in the test set; if two instances are classified into the same class, they are connected in the graph. Another graph $G^{T}$ is constructed for the target-domain instances as well by performing a clustering algorithm.
\item[2.] {\it Learner Weighting}: The weight of the $i$-th learner for the $j$-th target-domain instance $\mathbf{x}^{T}_j$ is proportional to the similarity between the instance's local structures in $G^{T}_{S_i}$ and $G^{T}$. And the similarity can be measured by the percentage of the common neighbors of $\mathbf{x}^{T}_j $ in these two graphs.
\end{itemize}
Note that this weighting scheme is based on the clustering-manifold assumption, i.e., if two instances are close to each other in a high-density region, they often have similar labels. In order to check the validity of this assumption for the task, the target task is tested on the source-domain training set(s). Specifically, the clustering quality of the training set(s) is quantified and checked by using a metric such as purity or entropy. If the clustering quality is not satisfactory, uniform weights are assigned to the learners instead. Besides, it is intuitive that if the measured structure similarity is particularly low for every learner, weighting and combining these learners seems unwise. Therefore, the authors introduce a threshold and compare it to the average similarity. If the similarity is lower than the threshold, the label of $\mathbf{x}^{T}_j$ is determined by the voting scheme among its reliable neighbors, where the reliable neighbors are the ones whose label predictions are made by the combined classifier.

The above-mentioned TaskTrAdaBoost and LWE approaches mainly focus on the ensemble process. In contrast, some studies focus more on the construction of weak learners. For example, Ensemble Framework of Anchor Adapters (ENCHOR) \cite{ZLP2016CIKM} is a weighting ensemble framework proposed by Zhuang {\it et al}. An anchor is a specific instance. Different from TrAdaBoost which adjusts weights of instances to train and produce a new learner iteratively, ENCHOR constructs a group of weak learners via using different representations of the instances produced by anchors. The thought is that the higher similarity between a certain instance and an anchor, the more likely the feature of that instance remains unchanged relative to the anchor, where the similarity can be measured by using the cosine or Gaussian distance function. ENCHOR contains the following steps.
\begin{itemize}
\item[1.] {\it Anchor Selection}: In this step, a group of anchors are selected. These anchors can be selected based on some rules or even randomly. In order to improve the final performance of ENCHOR, the authors proposed a method of selecting high-quality anchors \cite{ZLP2016CIKM}.

\item[2.] {\it Anchor-based Representation Generation}: For each anchor and each instance, the feature vector of an instance is directly multiplied by a coefficient that measures the distance from the instance to the anchor. In this way, each anchor produces a new pair of anchor-adapted source and target instance sets.

\item[3.] {\it Learner Training and Ensemble}: The obtained pairs of instance sets can be respectively used to train learners. Then, the resultant learners are weighted and combined to make the final predictions.

\end{itemize}
The framework ENCHOR is easy to be realized in a parallel manner in that the operations performed on each anchor are independent.

\subsection{Deep Learning Technique} \label{SEC.DL}
Deep learning methods are particularly popular in the field of machine learning. Many researchers utilize the deep learning techniques to construct transfer learning models. For example, the SDA and the mSLDA approaches mentioned in Section \ref{SEC.FE.EC} utilize the deep learning techniques. In this subsection, we specifically discuss the deep-learning-related transfer learning models. The deep learning approaches introduced are divided into two types, i.e., non-adversarial (or say, traditional) ones and adversarial ones.

\subsubsection{Traditional Deep Learning}
As said earlier, autoencoders are often used in deep learning area. In addition to SDA and mSLDA, there are some other reconstruction-based transfer learning approaches. For example, the paper by Zhuang {\it et al.} proposes an approach termed Transfer Learning with Deep Autoencoders (TLDA) \cite{ZCL2015IJCAI,ZCL2018TIST}. TLDA adopts two autoencoders for the source and the target domains, respectively. These two autoencoders share the same parameters. The encoder and the decoder both have two layers with activation functions. The diagram of the two autoencoders is presented as follows:
\begin{equation*}
\begin{aligned}
	&X^{S}{\xrightarrow{(W_1,b_1)}}Q^{S}{\xrightarrow[\text{Softmax Regression}]{(W_2,b_2)}}R^{S}{\xrightarrow{(\hat{W}_2,\hat{b}_2)}}\tilde{Q}^{S}{\xrightarrow{(\hat{W}_1,\hat{b}_1)}}\tilde{X}^{S},\\
	&\qquad\quad\overset{\Uparrow}{\underset{\Downarrow}{\text{\scriptsize KL Divergence}}}\\
	&X^{T}{\xrightarrow{(W_1,b_1)}}Q^{T}{\xrightarrow[\text{Softmax Regression}]{(W_2,b_2)}}R^{T}{\xrightarrow{(\hat{W}_2,\hat{b}_2)}}\tilde{Q}^{T}{\xrightarrow{(\hat{W}_1,\hat{b}_1)}}\tilde{X}^{T}.
\end{aligned}
\end{equation*}
There are several objectives of TLDA, which are listed as follows.
\begin{itemize}
\item[1.] {\it Reconstruction Error Minimization}: The output of the decoder should be extremely close to the input of encoder. In other words, the distance between $X^{S}$ and $\tilde{X}^{S}$ as well as the distance between $X^{T}$ and $\tilde{X}^{T}$ should be minimized.
\item[2.] {\it Distribution Adaptation}: The distribution difference between $Q^{S}$ and $Q^{T}$ should be minimized.
\item[3.] {\it Regression Error Minimization}: The output of the encoder on the labeled source-domain instances, i.e., $R^{S}$, should be consistent with the corresponding label information $Y^{S}$.
\end{itemize}
Therefore, the objective function of TLDA is given by
\begin{align*}
\min_{\Theta}~&{\mathcal{L}_{\text{REC}}(X,\tilde{X})+\lambda_{1}\text{KL}(Q^{S}||Q^{T})+\lambda_{2}\Omega(W,b,\hat{W},\hat{b})}\\&{+\lambda_{3}\mathcal{L}_{\text{REG}}(R^{S},Y^{S})},
\end{align*}
where the first term represents the reconstruction error, $\text{KL}(\cdot)$ represents the KL divergence, the third term controls the complexity, and the last term represents the regression error. TLDA is trained by using a gradient descent method. The final predictions can be made in two different ways. The first way is to directly use the output of the encoder to make predictions. And the second way is to treat the autoencoder as a feature extractor, and then train the target classifier on the labeled instances with the feature representation produced by the encoder's first-layer output.

In addition to the reconstruction-based domain adaptation, discrepancy-based domain adaptation is also a popular direction. In earlier research, the shallow neural networks are tried to learn the domain-independent feature representation \cite{GKZ2014PRICAI}. It is found that the shallow architectures often make it difficult for the resultant models to achieve excellent performance. Therefore, many studies turn to utilize deep neural networks. Tzeng {\it et al.} \cite{THZ2014ARXIV} added a single adaptation layer and a discrepancy loss to the deep neural network, which improves the performance. Further, Long {\it et al.} performed multi-layer adaptation and utilized multi-kernel technique, and they proposed an architecture termed Deep Adaptation Networks (DAN) \cite{LCW2015ICML}.

For better understanding, let us review DAN in detail. DAN is based on AlexNet \cite{KSH2012NIPS} and its architecture is presented below \cite{LCW2015ICML}.
\begin{equation*}
\begin{aligned}
	&&\hspace{-1em}\xrightarrow[\text{6th}]{\text{full}}R^{S}_6\xrightarrow[\text{7th}]{\text{full}}R^{S}_7\xrightarrow[\text{8th}]{\text{full}}R^{S}_{8}\left(f(X^{S})\right)\\
	\hspace{-0.5em}\myfrac{X^{S}}{X^{T}}&\underbrace{\xrightarrow[\text{1st}]{\text{conv}}\myfrac{Q^{S}_{1}}{Q^{T}_{1}}\xrightarrow[\cdots]{\text{conv}}\myfrac{Q^{S}_{5}}{Q^{T}_{5}}}&\hspace{-1em}\myfrac{^{\nearrow}}{_{\searrow}}\overset{\Uparrow}{\underset{\Downarrow}{\text{\scriptsize MK-MMD}}}\ \overset{\Uparrow}{\underset{\Downarrow}{\text{\scriptsize MK-MMD}}}\ \overset{\Uparrow}{\underset{\Downarrow}{\text{\scriptsize MK-MMD}}}\qquad\qquad\\
	&{\text{\scriptsize Five Convolutional Layers}}&\hspace{-1em}\underbrace{\xrightarrow[\text{6th}]{\text{full}}R^{T}_6\xrightarrow[\text{7th}]{\text{full}}R^{T}_7\xrightarrow[\text{8th}]{\text{full}}R^{T}_{8}\left(f(X^{T})\right)}\\
	&&{\text{\scriptsize Three Fully Connected Layers}\qquad\qquad}
\end{aligned}
\end{equation*}
In the above network, the features are first extracted by five convolutional layers in a general-to-specific manner. Next, the extracted features are fed into one of the two fully connected networks switched by their original domains. These two networks both consist of three fully connected layers that are specialized for the source and the target domains. DAN has the following objectives.
\begin{itemize}
\item[1.] {\it Classification Error Minimization}: The classification error of the labeled instances should be minimized. The cross-entropy loss function is adopted to measure the prediction error of the labeled instances.

\item[2.] {\it Distribution Adaptation}: Multiple layers, which include the representation layers and the output layer, can be jointly adapted in a layer-wise manner. Instead of using the single-kernel MMD to measure the distribution difference, the authors turn to MK-MMD. The authors adopt the linear-time unbiased estimation of MK-MMD to avoid numerous inner product operations \cite{GSS2012NIPS}.

\item[3.] {\it Kernel Parameter Optimization}: The weighting parameters of the multiple kernels in MK-MMD should be optimized to maximize the test power \cite{GSS2012NIPS}.
\end{itemize}
The objective function of the DAN network is given by:
\begin{equation*}
\min_{\Theta}\max_{\kappa}{\sum_{i=1}^{n^{L}}\mathcal{L}\left(f(\mathbf{x}^{L}_{i}),y^{L}_{i}\right)+\lambda\sum_{l=6}^{8}\text{MK-MMD}(R^{S}_{l},R^{T}_{l};\kappa)},
\end{equation*}
where $l$ denotes the index of the layer. The above optimization is actually a minimax optimization problem. The maximization of the objective function with respect to the kernel function $\kappa$ aims to maximize the test power. After this step, the subtle difference between the source and the target domains are magnified. This train of thought is similar to the Generative Adversarial Network (GAN) \cite{GPM2014NIPS}. In the training process, the DAN network is initialized by a pre-trained AlexNet \cite{KSH2012NIPS}. There are two categories of parameters that should be learned, i.e., the network parameters and the weighting parameters of the multiple kernels. Given that the first three convolutional layers output the general features and are transferable, the authors freeze them and fine-turn the last two convolutional layers and the two fully connected layers \cite{YCB2014NIPS}. The last fully connected layer (or say, the classifier layer) is trained from scratch.

Long {\it et al.} further extended the above DAN approach and proposed the DAN framework \cite{LCC20XXTPAMI}. The new characteristics are summarized as follows.
\begin{itemize}
\item[1.] {\it Regularizer Adding}: The framework introduces an additional regularizer to minimize the uncertainty of the predicted labels of the unlabeled target-domain instances, which is motivated by entropy minimization criterion \cite{GB2004NIPS}.

\item[2.] {\it Architecture Generalizing}: The DAN framework can be applied to many other architectures such as GoogLeNet \cite{SLJ2015CVPR} and ResNet \cite{HZR2016CVPR}.

\item[3.] {\it Measurement Generalizing}: The distribution difference can be estimated by other metrics. For example, in addition to MK-MMD, the authors also present the Mean Embedding test for distribution adaptation \cite{CRS2015NIPS}.
\end{itemize}
The objective function of the DAN framework is given by:
\begin{equation*}
\begin{aligned}
	\min_{\Theta}\max_{\kappa}&{\sum_{i=1}^{n^{L}}\mathcal{L}\left(f(\mathbf{x}^{L}_{i}),y^{L}_{i}\right)+\lambda_{1}\sum_{l=l_{\text{strt}}}^{l_{\text{end}}}\text{DIST}(R^{S}_{l},R^{T}_{l})}\\
	&+\lambda_{2}\sum_{i=1}^{n^{T,U}}\sum_{y_j\in\mathcal{Y}}{S}\left(P(y_j|f(\mathbf{x}^{T,U}_{i}))\right),
\end{aligned}
\end{equation*}
where $l_{\text{strt}}$ and $l_{\text{end}}$ denote the boundary indexes of the fully connected layers for adapting the distributions.

There are some other impressive works. For example, Long {\it et al.} constructed residual transfer networks for domain adaptation, which is motivated by deep residual learning \cite{LZW2016NIPS}. Besides, another work by Long {\it et al.} proposes the Joint Adaptation Network (JAN) \cite{LZW2017ICML}, which adapts the joint distribution difference of multiple layers. Sun and Saenko extended CORAL for deep domain adaptation and proposed an approach termed Deep CORAL (DCORAL), in which the CORAL loss is added to minimize the feature covariance \cite{SS2016ECCVW}. Chen {\it et al.} realized that the instances with the same label should be close to each other in the feature space, and they not only add the CORAL loss but also add an instance-based class-level discrepancy loss \cite{CCJ2019AAAI}. Pan {\it et al.} constructed three prototypical networks (corresponding to $\mathcal{D}_{S},\mathcal{D}_{T}$ and $\mathcal{D}_{S}\cup\mathcal{D}_{T}$) and incorporated the thought of multi-model consensus. They also adopt pseudo-label strategy and adapt both the instance-level and class-level discrepancy \cite{PYL2019CVPR}. Kang {\it et al.} proposed the Contrastive Adaptation Network (CAN), which is based on the discrepancy metric termed contrastive domain discrepancy \cite{KJY2019CVPR}. Zhu {\it et al.} aimed to adapt the extracted multiple feature representations and proposed the Multi-Representation Adaptation Network (MRAN) \cite{ZZW2019NN}.

Deep learning technique can also be used for multi-source transfer learning. For example, the work by Zhu {\it et al.} proposes a framework that is referred to as Multiple Feature Spaces Adaptation Network (MFSAN) \cite{ZZW2019AAAI}. The architecture of MFSAN consists of a common-feature extractor, $m^{S}$ domain-specific feature extractors, and $m^{S}$ domain-specific classifiers. The corresponding schematic diagram is shown below.
\begin{align*}
&\myfrac{X_{1}^{S}\cdots X_{k}^{S}\cdots X_{m^{S}}^{S}}{X^{T}}{\xrightarrow[\text{Extractor}]{\text{Common}}}\myfrac{Q_{1}^{S}\cdots Q_{k}^{S}\cdots Q_{m^{S}}^{S}}{Q^{T}}{\xrightarrow[\text{ Extractors}]{\text{Domain-Specific}}}\\&\myfrac{R_{1}^{S}\cdots R_{k}^{S}\cdots R_{m^{S}}^{S}}{R_{1}^{T}\cdots R_{k}^{T}\cdots R_{m^{S}}^{T}}{\xrightarrow[\text{Classifiers}]{\text{Domain-Specific}}}\myfrac{\hat{Y}_{1}^{S}\cdots\hat{Y}_{k}^{S}\cdots\hat{Y}_{m^{S}}^{S}}{\hat{Y}_{1}^{T}\cdots\hat{Y}_{k}^{T}\cdots\hat{Y}_{m^{S}}^{T}}
\end{align*}
In each iteration, MFSAN has the following steps.
\begin{itemize}
\item[1.] {\it Common Feature Extraction}: For each source domain (denoted by $\mathcal{D}_{S_k}$ with $k=1,\cdots,m^{S}$), the source-domain instances (denoted by $X_{k}^{S}$) are separately input to the common-feature extractor to produce instances in a common latent feature space (denoted by $Q_{k}^{S}$). Similar operations are also performed on the target-domain instances (denoted by $X^{T}$), which produces $Q^{T}$.
\item[2.] {\it Specific Feature Extraction}: For each source domain, the extracted common features $Q_{k}^{S}$ is fed to the $k$-th domain-specific feature extractor. Meanwhile, $Q^{T}$ is fed to all the domain-specific feature extractors, which results in $R^{T}_{k}$ with $k=1,\cdots,m^{S}$.
\item[3.] {\it Data Classification}: The output of the $k$-th domain-specific feature extractor is input to the $k$-th classifier. In this way, $m^{S}$ pairs of the classification results are predicted in the form of probability.
\item[4.] {\it Parameter Updating}: The parameters of the network are updated to optimize the objective function.
\end{itemize}
There are three objectives in MFSAN, i.e., classification error minimization, distribution adaptation, and consensus regularization. The objective function is given by:
\begin{equation*}
\begin{aligned}
	\min_{\Theta}~&{\sum_{i=1}^{m^{S}}\mathcal{L}(\hat{Y}_{i}^{S},Y_{i}^{S})+\lambda_{1}\sum_{i=1}^{m^{S}}\text{MMD}(R_{i}^{S},R_{i}^{T})}\\&+\lambda_{2}\sum_{i\ne j}^{m^{S}}\left\vert\hat{Y}_{i}^{T}-\hat{Y}_{j}^{T}\right\vert,
\end{aligned}
\end{equation*}
where the first term represents the classification error of the labeled source-domain instances, the second term measures the distribution difference, and the third term measures the discrepancy of the predictions on the target-domain instances.

\subsubsection{Adversarial Deep Learning}
The thought of adversarial learning can be integrated into deep-learning-based transfer learning approaches. As mentioned above, in the DAN framework, the network $\Theta$ and the kernel $\kappa$ play a minimax game, which reflects the thought of adversarial learning. However, the DAN framework is a little different from the traditional GAN-based methods in terms of the adversarial matching. In the DAN framework, there is only a few parameters to be optimized in the max game, which makes the optimization easier to achieve equilibrium. Before introducing the adversarial transfer learning approaches, let us briefly review the original GAN framework and the related work.

The original GAN \cite{GPM2014NIPS}, which is inspired by the two-player game, is composed of two models, a generator $\mathpzc{G}$ and a discriminator $\mathpzc{D}$. The generator produces the counterfeits of the true data for the purpose of confusing the discriminator and making the discriminator produce wrong detection. The discriminator is fed with the mixture of the true data and the counterfeits, and it aims to detect whether a data is the true one or the fake one. These two models actually play a two-player minimax game, and the objective function is as follows:
\begin{equation*}
\min_{\mathpzc{G}}\max_{\mathpzc{D}}{\mathbb{E}_{\mathbf{x}\sim P_{\text{true}}}\left[\log\mathpzc{D}(\mathbf{x})\right]+\mathbb{E}_{\tilde{\mathbf{z}}\sim P_{\tilde{\mathbf{z}}}}\left[\log\left(1-\mathpzc{D}(\mathpzc{G}(\tilde{\mathbf{z}}))\right)\right]},
\end{equation*}
where $\tilde{\mathbf{z}}$ represents the noise instances (sampled from a certain noise distribution) used as the input of the generator for producing the counterfeits. The entire GAN can be trained by using the back-propagation algorithm. When the two-player game achieves equilibrium, the generator can produce almost true-looking instances.

Motivated by GAN, many transfer learning approaches are established based on the assumption that a good feature representation contains almost no discriminative information about the instances' original domains. For example, the work by Ganin {\it et al.} proposes a deep architecture termed Domain-Adversarial Neural Network (DANN) for domain adaptation \cite{GL2015ICML,GUA2016JMLR}. DANN assumes that there is no labeled target-domain instance to work with. Its architecture consists of a feature extractor, a label predictor, and a domain classifier. The corresponding diagram is as follows.
\begin{equation*}
\begin{aligned}
	&\hspace{-2.5em}\xrightarrow[\text{Predictor}]{\text{Label}}\myfrac{\hat{Y}^{S,L}}{\hat{Y}^{T,U}}\\
	\myfrac{X^{S,L}}{X^{T,U}}\xrightarrow[\text{Extractor}]{\text{Feature}}\myfrac{\overset{\uparrow}{Q^{S,L}}}{Q^{T,U}}\bigg\rbrace&\xrightarrow[\text{Classifier}]{\text{Domain}}\myfrac{\hat{S}}{\hat{T}}~(\text{\scriptsize Domain Label})
\end{aligned}
\end{equation*}
The feature extractor acts like the generator, which aims to produce the domain-independent feature representation for confusing the domain classifier. The domain classifier plays the role like the discriminator, which attempts to detect whether the extracted features come from the source domain or the target domain. Besides, the label predictor produces the label prediction of the instances, which is trained on the extracted features of the labeled source-domain instances, i.e., $Q^{S,L}$. DANN can be trained by inserting a special gradient reversal layer (GRL). After the training of the whole system, the feature extractor learns the deep feature of the instances, and the output $\hat{Y}^{T,U}$ is the predicted labels of the unlabeled target-domain instances.

There are some other related impressive works. The work by Tzeng {\it et al.} proposes a unified adversarial domain adaptation framework \cite{THS2017CVPR}. The work by Shen {\it et al.} adopts Wasserstein distance for domain adaptation \cite{SQZ2018AAAI}. Hoffman {\it et al.} adopted cycle-consistency loss to ensure the structural and semantic consistency \cite{HTP2018ICML}. Long {\it et al.} proposed the Conditional Domain Adversarial Network (CDAN), which utilizes a conditional domain discriminator to assist adversarial adaptation \cite{LCW2018NIPS}. Zhang {\it et al.} adopted a symmetric design for the source and the target classifiers \cite{ZTJ2019CVPR}. Zhao {\it et al.} utilized domain adversarial networks to solve the multi-source transfer learning problem \cite{ZZW2018NIPS}. Yu {\it et al.} proposed a dynamic adversarial adaptation network \cite{YWC2019ICDM}.

Some approaches are designed for some special scenarios. Take the partial transfer learning as an example. The partial transfer learning approaches are designed for the scenario that the target-domain classes are less than the source-domain classes, i.e., $\mathcal{Y}^{S}\subseteq\mathcal{Y}^{T}$. In this case, the source-domain instances with different labels may have different importance for domain adaptation. To be more specific, the source-domain and the target-domain instances with the same label are more likely to be potentially associated. However, since the target-domain instances are unlabeled, how to identify and partially transfer the important information from the labeled source-domain instances is a critical issue.

The paper by Zhang {\it et al.} proposes an approach for partial domain adaptation, which is called Importance Weighted Adversarial Nets-Based Domain Adaptation (IWANDA) \cite{ZDL2018CVPR}. The architecture of IWANDA is different from that of DANN. DANN adopts one common feature extractor based on the assumption that there exists a common feature space where $Q^{S,L}$ and $Q^{T,U}$ have the similar distribution. However, IWANDA uses two domain-specific feature extractors for the source and the target domains, respectively. Specifically, IWANDA consists of two feature extractors, two domain classifiers, and one label predictor. The diagram of IWANDA is presented below.
\begin{equation*}
\begin{aligned}
	&\hspace{-2.5em}\xrightarrow[\text{Predictor}]{\text{Label}}\myfrac{\hat{Y}^{S,L}}{\hat{Y}^{T,U}}\\
	\myfrac{X^{S,L}\xrightarrow[\text{Extractor}]{\text{Source Feature}}\overset{\uparrow}{Q^{S,L}}}{X^{T,U}\xrightarrow[\text{Extractor}]{\text{Target Feature}}\underset{\downarrow}{Q^{T,U}}}\bigg\rbrace&\xrightarrow[+\hat{Y}^{T,U}]{+\bm\beta^{S}}\xrightarrow[\text{Classifier}]{\text{2nd Domain}}\myfrac{\hat{S_2}}{\hat{T_2}}\\
	&\hspace{-2.5em}\xrightarrow[\text{Classifier}]{\text{1st Domain}}\myfrac{\hat{S_1}}{\hat{T_1}}\xrightarrow[\text{Function}]{\text{Weight}}\myfrac{\bm\beta^{S}}{~}
\end{aligned}
\end{equation*}
Before training, the source feature extractor and the label predictor are pre-trained on the labeled source-domain instances. These two components are frozen in the training process, which means that only the target feature extractor and the domain classifiers should be optimized. In each iteration, the above network is optimized by taking the following steps.
\begin{itemize}	
\item[1.] {\it Instance Weighting}: In order to solve the partial transfer issue, the source-domain instances are assigned with weights based on the output of the first domain classifier. The first domain classifier is fed with $Q^{S,L}$ and $Q^{T,U}$, and then outputs the probabilistic predictions of their domains. If a source domain instance is predicted with a high probability of belonging to the target domain, this instance is highly likely to associate with the target domain. Thus, this instance is assigned with a larger weight and vice versa.

\item[2.] {\it Prediction Making}: The label predictor outputs the label predictions of the instances. The second classifier predicts which domain an instance belongs to.

\item[3.] {\it Parameter Updating}: The first classifier is optimized to minimize the domain classification error. The second classifier plays a minmax game with the target feature extractor. This classifier aims to detect whether a instance is the instance from the target domain or the weighted instance from the source domain, and to reduce the uncertainty of the label prediction $\hat{Y}^{T,U}$. The target feature extractor aims to confuse the second classifier. These components can be optimized in a similar way to GAN or by inserting a GRL.
\end{itemize}

In addition to IWANDA, the work by Cao {\it et al.} constructs the selective adversarial network for partial transfer learning \cite{CLW2018CVPR}. There are some other studies related to transfer learning. For example, the work by Wang {\it et al.} proposes a minimax-based approach to select high-quality source-domain data \cite{WQW2019KDD}. Chen {\it et al.} investigated the transferability and the discriminability in the adversarial domain adaptation, and proposed a spectral penalization approach to boost the existing adversarial transfer learning methods \cite{CWL2019ICML}.

\section{Application} \label{SEC.AC}
In previous sections, a number of representative transfer learning approaches are introduced, which have been applied to solving a variety of text-related/image-related problems in their original papers. For example, MTrick \cite{ZLX2011SADM} and TriTL \cite{ZLD2014TC} utilize the matrix factorization technique to solve cross-domain text classification problems; the deep-learning-based approaches such as DAN \cite{LCW2015ICML}, DCORAL \cite{SS2016ECCVW}, and DANN \cite{GL2015ICML,GUA2016JMLR} are applied to solving image classification problems. Instead of focusing on the general text-related or image-related applications, in this section, we mainly focus on the transfer learning applications in specific areas such as medicine, bioinformatics, transportation, and recommender systems.

\subsection{Medical Application}
Medical imaging plays an important role in the medical area, which is a powerful tool for diagnosis. With the development of computer technology such as machine learning, computer-aided diagnosis has become a popular and promising direction. Note that medical images are generated by special medical equipment, and their labeling often relies on experienced doctors. Therefore, in many cases, it is expensive and hard to collect sufficient training data. Transfer learning technology can be utilized for medical imaging analysis. A commonly used transfer learning approach is to pre-train a neural network on the source domain (e.g., ImageNet, which is an image database containing more than fourteen million annotated images with more than twenty thousand categories \cite{DDS2009CVPR}) and then finetune it based on the instances from the target domain.

For example, Maqsood {\it et al.} finetuned the AlexNet \cite{KSH2012NIPS} for the detection of Alzheimer's disease \cite{MNK2019S}. Their approach has the following four steps. First, the MRI images from the target domain are pre-processed by performing contrast stretching operations. Second, the AlexNet architecture \cite{KSH2012NIPS} is pre-trained over ImageNet \cite{DDS2009CVPR} (i.e., the source domain) as a starting point to learn the new task. Third, the convolutional layers of AlexNet are fixed, and the last three fully connected layers are replaced by the new ones including one softmax layer, one fully connected layer, and one output layer. Finally, the modified AlexNet is finetuned by training on the Alzheimer's dataset \cite{MFC2010JCN} (i.e., the target domain). The experimental results show that the proposed approach achieves the highest accuracy for the multi-class classification problem (i.e, Alzheimer's stage detection).

Similarly, Shin {\it et al.} finetuned the pre-trained deep neural network for solving the computer-aided detection problems \cite{SRG2016TMI}. Byra {\it et al.} utilized the transfer learning technology to help assess knee osteoarthritis \cite{BWZ2019MRM}. In addition to imaging analysis, transfer learning has some other applications in the medical area. For example, the work by Tang {\it et al.} combines the active learning and the domain adaptation technologies for the classification of various medical data \cite{TDH2019ICV}. Zeng {\it et al.} utilized transfer learning for automatically encoding ICD-9 codes that are used to describe a patient's diagnosis \cite{ZLF2019NC}.

\subsection{Bioinformatics Application}
Biological sequence analysis is an important task in the bioinformatics area. Since the understanding of some organisms can be transferred to other organisms, transfer learning can be applied to facilitate the biological sequence analysis. The distribution difference problem exists significantly in this application. For example, the function of some biological substances may remain unchanged but with the composition changed between two organisms, which may result in the marginal distribution difference. Besides, if two organisms have a common ancestor but with long evolutionary distance, the conditional distribution difference would be significant. The work by Schweikert {\it et al.} uses the mRNA splice site prediction problem as the example to analyze the effectiveness of transfer learning approaches \cite{SRW2008NIPS}. In their experiments, the source domain contains the sequence instances from a well-studied model organism, i.e., {\it C. elegans}, and the target organisms include two additional nematodes (i.e., {\it C. remanei} and {\it P. pacificus}), {\it D. melanogaster}, and the plant {\it A. thaliana}. A number of transfer learning approaches, e.g., FAM \cite{D2007ACL} and the variant of KMM \cite{HSG2006NIPS}, are compared with each other. The experimental results show that transfer learning can help improve the classification performance.
	
Another widely encountered task in the bioinformatics area is gene expression analysis, e.g., predicting associations between genes and phenotypes. In this application, one of the main challenges is the data sparsity problem, since there is usually very little data of the known associations. Transfer learning can be used to leverage this problem by providing additional information and knowledge. For example, Petegrosso {\it et al.} \cite{PPHK2017B} proposed a transfer learning approach to analyze and predict the gene-phenotype associations based on the Label Propagation Algorithm (LPA) \cite{HK2010SICDM}. LPA utilizes the Protein-Protein Interaction (PPI) network and the initial labeling to predict the target associations based on the assumption that the genes that are connected in the PPI network should have the similar labels. The authors extended LPA by incorporating multi-task and transfer-learning technologies. First, Human Phenotype Ontology (HPO), which provides a standardized vocabulary of phenotypic features of human diseases, is utilized to form the auxiliary task. In this way, the associations can be predicted by utilizing phenotype paths and both the linkage knowledge in HPO and in the PPI network; the interacted genes in PPI are more likely to be associated with the same phenotype and the connected phenotypes in HPO are more likely to be associated with the same gene. Second, Gene Ontology (GO), which contains the association information between gene functions and genes, is used as the source domain. Additional regularizers are designed, and the PPI network and the common genes are used as the bridge for knowledge transfer. The gene-GO term and gene-HPO phenotype associations are constructed simultaneously for all the genes in the PPI network. By transferring additional knowledge, the predicted gene-phenotype associations can be more reliable.
	
Transfer learning can also be applied to solving the PPI prediction problems. Xu {\it et al.} \cite{XXY2010ICBB} proposed an approach to transfer the linkage knowledge from the source PPI network to the target one. The proposed approach is based on the collective matrix factorization technique \cite{SG2008KDD}, where a factor matrix is shared across domains.

\subsection{Transportation Application}
One application of transfer learning in the transportation area is to understand the traffic scene images. In this application, a challenge problem is that the images taken from a certain location often suffer from variations because of different weather and light conditions. In order to solve this problem, Di {\it et al.} proposed an approach that attempts to transfer the information of the images that were taken from the same location in different conditions \cite{DZL2018TITS}. In the first step, a pre-trained network is finetuned to extract the feature representations of images. In the second step, the feature transformation strategy is adopted to construct a new feature representation. Specifically, the dimension reduction algorithm (i.e., partial least squares regression \cite{A2010WIRCS}) is performed on the extracted features to generate low-dimension features. Then, a transformation matrix is learned to minimize the domain discrepancy of the dimension-reduced data. Next, the subspace alignment operations are adopted to further reduce the domain discrepancy. Note that, although images under different conditions often have different appearances, they often have the similar layout structure. Therefore, in the final step, the cross-domain dense correspondences are established between the test image and the retrieved best matching image at first, and then the annotations of the best matching image are transferred to the test image via the Markov random field model \cite{PPL1997TPAMI,LYT2011PAMI}.

Transfer learning can also be applied to the task of driver behavior modeling. In this task, sufficient personalized data of each individual driver are usually unavailable. In such situations, transferring the knowledge contained in the historical data for the newly-involved driver is a promising alternative. For example, Lu {\it et al.} proposed an approach to driver model adaptation in lane-changing scenarios \cite{LHC2019TITS}. The source domain contains the sufficient data describing the behavior of the source drivers, while the target domain has a few numbers of data about the target driver. In the first step, the data from both domains are pre-processed by performing PCA to generate low-dimension features. The authors assume that the source and the target data are from two manifolds. Therefore, in the second step, a manifold alignment approach is adopted for domain adaptation. Specifically, the dynamic time warping algorithm \cite{BC1994KDD} is applied to measuring similarity and finding the corresponding source-domain data point of each target-domain data point. Then, local Procrustes analysis \cite{MHR2018JRM} is adopted to align the two manifolds based on the obtained correspondences between data points. In this way, the data from the source domain can be transferred to the target domain. And in the final step, a stochastic modeling method (e.g., Gaussian mixture regression \cite{AMT2011IIVS}) is used to model the behavior of the target driver. The experimental results demonstrate that the transfer learning approach can help the target driver even when few target-domain data are available. Besides, the results also show that when the number of target instances are very small or very large, the superiority of their approach is not obvious. This may because the relationship across domains cannot be found exactly with few target-domain instances, and in the case of sufficient target-domain instances, the necessity of transfer learning is reduced.

Besides, there are some other applications of transfer learning in the transportation area. For example, Liu {\it et al.} applied transfer learning to driver poses recognition \cite{LLP2019TITS}. Wang {\it et al.} adopted the regularization technique in transfer learning for vehicle type recognition \cite{WZH2018TITS}. Transfer learning can also be utilized for anomalous activity detection \cite{GKC2017CBM,BN2019JIFS}, traffic sign recognition \cite{RSZ2018IRI}, etc.

\subsection{Recommender-System Application}
Due to the rapid increase of the amount of information, how to effectively recommend the personalized content for individual users is an important issue. In the field of recommender systems, some traditional recommendation methods, e.g., factorization-based collaborative filtering, often rely on the factorization of the user-item interaction matrix to obtain the predictive function. These methods often require a large amount of training data to make accurate recommendations. However, the necessary training data, e.g., the historical interaction data, are often sparse in real-world scenarios. Besides, for new registered users or new items, traditional methods are often hard to make effective recommendations, which is also known as the cold-start problem.

Recognizing the above-mentioned problems in recommender systems, kinds of transfer learning approaches, e.g., instance-based and feature-based approaches, have been proposed. These approaches attempt to make use of the data from other recommender systems (i.e., the source domains) to help construct the recommender system in the target domain. Instance-based approaches mainly focus on transferring different types of instances, e.g., ratings, feedbacks, and examinations, from the source domain to the target domain. The work by Pan {\it et al.} \cite{PXY2012AAAI} leverages the uncertain ratings (represented as rating distributions) of the source domain for knowledge transfer. Specifically, the source-domain uncertain ratings are used as constraints to help complete the rating matrix factorization task on the target domain. Hu {\it et al.} \cite{HZY2019WWW} proposed an approach termed transfer meeting hybrid, which extracts the knowledge from unstructured text by using an attentive memory network and selectively transfer the useful information.

Feature-based approaches often leverage and transfer the information from a latent feature space. For example, Pan {\it et al.} proposed an approach termed Coordinate System Transfer (CST) \cite{PXL2010AAAI} to leverage both the user-side and the item-side latent features. The source-domain instances come from another recommender system, sharing common users and items with the target domain. CST is developed based on the assumption that the principle coordinates, which reflect the tastes of users or the factors of items, characterize the domain-independent structure and are transferable across domains. CST first constructs two principle coordinate systems, which are actually the latent features of users and items, by applying sparse matrix tri-factorization on the source-domain data, and then transfer the coordinate systems to the target domain by setting them as constraints. The experimental results show that CST significantly outperforms the non-transfer baselines (i.e., average filling model and latent factorization model) in all data sparsity levels \cite{PXL2010AAAI}.

There are some other studies about cross-domain recommendation \cite{PY2013AI,ZZZ2017WWWC,ZZS2017CIKM,ZZC2018NN}. For example, He {\it et al.} proposed a transfer learning framework based on the Bayesian neural network \cite{HLZ2018ICDM}. Zhu {\it et al.} \cite{ZWC2018IJCAI} proposed a deep framework, which first generates the user and item feature representations based on the matrix factorization technique, and then employs a deep neural network to learn the mapping of features across domains. Yuan {\it et al.} \cite{YYB2019IJCAI} proposed a deep domain adaptation approach based on autoencoders and a modified DANN \cite{GL2015ICML,GUA2016JMLR} to extract and transfer the instances from rating matrices.

\subsection{Other Applications}
\noindent{\textbf{Communication Application}}: In addition to WiFi localization tasks \cite{PY2010TKDE,PTK2011TNN}, transfer learning has also been employed in wireless-network applications. For example, Bastug {\it et al.} proposed a caching mechanism \cite{BBD2015WIOPT}; the knowledge contained in contextual information, which is extracted from the interactions between devices, is transferred to the target domain. Besides, some studies focus on the energy saving problems. The work by Li {\it et al.} proposes an energy saving scheme for cellular radio access networks, which utilizes the transfer-learning expertise \cite{LZC2014TWC}. The work by Zhao and Grace applies transfer learning to topology management for reducing energy consumption \cite{ZG2014ICUC}.

\noindent{\textbf{Urban-Computing Application}}: With a large amount of data related to our cities, urban-computing is a promising researching track in directions of traffic monitoring, health care, social security, etc. Transfer learning has been applied to alleviate the data scarcity problem in many urban computing applications. For example, Guo {\it et al.} \cite{GLZ2018IMWUT} proposed an approach for chain store site recommendation, which leverages the knowledge from semantically-relevant domains (e.g., other cities with the same store and other chain stores in the target city) to the target city. Wei {\it et al.} \cite{WZY2016KDD} proposed a flexible multi-modal transfer learning approach that transfers knowledge from a city that have sufficient multi-model data and labels to the target city to alleviate the data sparsity problem.

Transfer learning has been applied to some recognition tasks such as hand gesture recognition \cite{CFD2019TNSRE}, face recognition \cite{RDH2014TIP}, activity recognition \cite{WCH2018ICPCC}, and speech emotion recognition \cite{DZM2013ACII}. Besides, transfer-learning expertise has also been incorporated into some other areas such as sentiment analysis \cite{LSJ2019ACCESS,GBB2011ICML,XZZ2020WWW}, fraud detection \cite{ZXS2020WWW}, social network \cite{TLK2016TIS}, and hyperspectral image analysis \cite{SZL2019ICV,ZZT2014TGRS}.

\section{Experiment} \label{SEC.EX}
Transfer learning techniques have been successfully applied in many real-world applications. In this section, we perform experiments to evaluate the performance of some representative transfer learning models\footnote{https://github.com/FuzhenZhuang/Transfer-Learning-Toolkit}~\cite{ZDG2019ARXIV} of different categories on two mainstream research areas, i.e., object recognition and text classification. The datasets are introduced at first. Then, the experimental results and further analyses are provided.

\subsection{Dataset and Preprocessing}
\begin{table*}[!t]
\centering
\caption{Statistical information of the preprocessed datasets.}
\setlength{\tabcolsep}{1.8mm}{
	\begin{tabular}{cccccc}
		\toprule
		Area & Dataset & Domain & Attribute & Total Instances & Tasks \\
		\midrule
		Sentiment Classification & Amazon Reviews &  4   &   5000     & 27677 & 12 \\
		\midrule
		Text Classification & Reuters-21578 &  3   &   4772  &  6570   & 3 \\
		\midrule
		Object Recognition & Office-31 &   3  &  800   &   4110  & 6 \\
		\bottomrule
	\end{tabular}%
}
\label{tab:dataset}%
\end{table*}%
Three datasets are studied in the experiments, i.e., Office-31, Reuters-21578, and Amazon Reviews. For simplicity, we focus on the classification tasks. The statistical information of the preprocessed datasets is listed in Table \ref{tab:dataset}.

\begin{itemize}[leftmargin=*]
\item \textbf{Amazon Reviews}\footnote{http://www.cs.jhu.edu/~mdredze/datasets/sentiment/}~\cite{BDP2007ACL} is a multi-domain sentiment dataset which contains product reviews taken from Amazon.com of four domains (Books, Kitchen, Electronics and DVDs). Each review in the four domains has a text and a rating from zero to five. In the experiments, the ratios that are less than three are defined as the negative ones, while others are defined as the positive ones. The frequency of each word in all reviews is calculated. Then, the five thousand words with the highest frequency are selected as the attributes of each review. In this way, we finally have 1000 positive instances, 1000 negative instances, and about 5000 unlabeled instances in each domain. In the experiments, every two of the four domains are selected to generate twelve tasks.

\item \textbf{Reuters-21578}\footnote{https://archive.ics.uci.edu/ml/datasets/Reuters-21578+Text+Categorization+Collection} is a dataset for text categorization, which has a hierarchical structure. The dataset contains 5 top categories (Exchanges, Orgs, People, Places, Topics). In out experiment, we use the top three big category Orgs, People and Places to generate three classification tasks (Orgs vs People, Orgs vs Places and People vs Places). In each task, the subcategories in the corresponding two categories are separately divided into two parts. Then, the resultant four parts are used as the components to form two domains. Each domain has about 1000 instances, and each instance has about 4500 features. Specifically, taking the task Orgs vs People as an example, one part from Orgs and one part from People and combined to form the source domain; similarly, the rest two parts form the target domain. Note that the instances in the three categories are all labeled. In order to generate the unlabeled instances, the labeled instances are selected from the dataset, and their labels are ignored.

\item \textbf{Office-31} \cite{SKF2010ECCV} is an object recognition dataset which contains thirty-one categories and three domains, i.e., Amazon, Webcam, and DSLR. These three domains have 2817, 498, and 795 instances, respectively. The images in Amazon are the online e-commerce pictures taken from Amazon.com. The images in Webcam are the low-resolution pictures taken by web cameras. And the images in DSLR are the high-resolution pictures taken by DSLR cameras. In the experiments, every two of the three domains (with the order considered) are selected as the source and the target domains, which results in six tasks.
\end{itemize}

\subsection{Experiment Setting}
Experiments are conducted to compare some representative transfer learning models. Specifically, eight algorithms are performed on the dataset Office-31 for solving the object recognition problem. Besides, fourteen algorithms are performed and evaluated on the dataset Reuters-21578 for solving the text classification problem. In the sentiment classification problem, eleven algorithms are performed on Amazon Reviews. The classification results are evaluated by accuracy, which is defined as follows:
\begin{equation*}
\text{accuracy} = \frac{|\{ \mathbf{x} | \mathbf{x}_{i} \in \mathcal{D}_{\text{test}} \land f(\mathbf{x}_{i})=y_{i}  \}|}{|\mathcal{D}_{\text{test}}|}
\end{equation*}
where $\mathcal{D}_{\text{test}}$ denotes the test data and $y$ denotes the truth classification label; $f(\mathbf{x})$ represents the predicted classification result. Note that some algorithms need the base classifier. In these cases, an SVM with a linear kernel is adopted as the base classifier in the experiments. Besides, the source-domain instances are all labeled. And for the performed algorithms (except TrAdaBoost), the target-domain instances are unlabeled. Each algorithm was executed three times, and the average results are adopted as our experimental results.

The evaluated transfer learning models include: HIDC \cite{ZLY2013IJCAI}, TriTL \cite{ZLD2014TC}, CD-PLSA \cite{ZLS2010CIKM,ZLS2012TKDE}, MTrick \cite{ZLX2011SADM}, SFA \cite{PNS2010WWW}, mSLDA \cite{CXW2012ICML,CWX2015JMLR}, SDA \cite{GBB2011ICML}, GFK \cite{G2012CVPR}, SCL \cite{BMP2006EMNLP}, TCA \cite{PTK2009IJCAI,PTK2011TNN}, CoCC \cite{DXY2007KDD}, JDA \cite{LWD2013ICCV}, TrAdaBoost \cite{DYX2007ICML}, DAN \cite{LCW2015ICML}, DCORAL \cite{SS2016ECCVW}, MRAN \cite{ZZW2019NN}, CDAN \cite{LCW2018NIPS}, DANN \cite{GL2015ICML,GUA2016JMLR}, JAN \cite{LZW2017ICML}, and CAN \cite{KJY2019CVPR}.

\subsection{Experiment Result}
\begin{figure}
\centering
\includegraphics[width=0.9\linewidth]{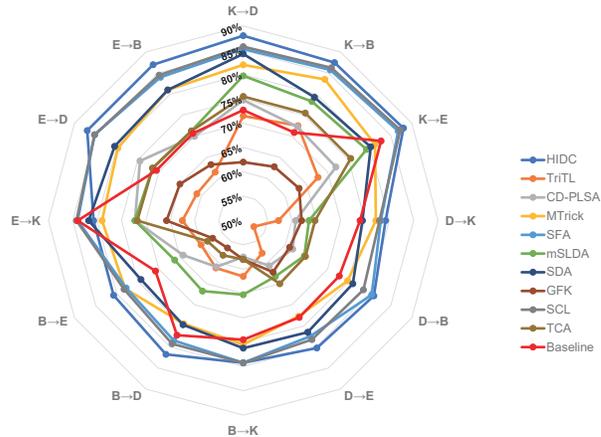}
\caption{Comparison results on Amazon Reviews.}
\label{fig:Amazon_results}
\end{figure}

\begin{figure*}
\centering
\includegraphics[width=0.49\linewidth]{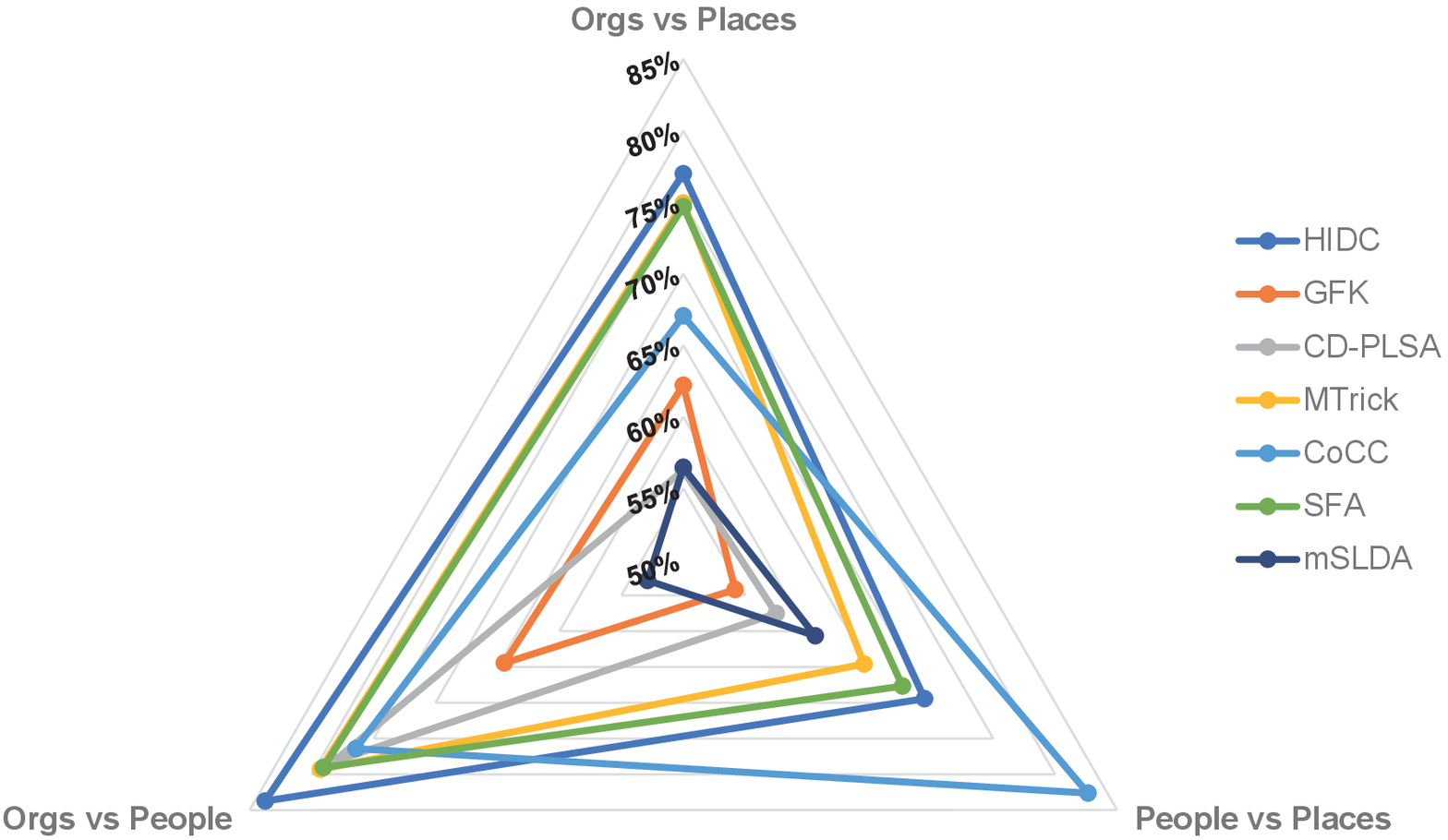}
\includegraphics[width=0.49\linewidth]{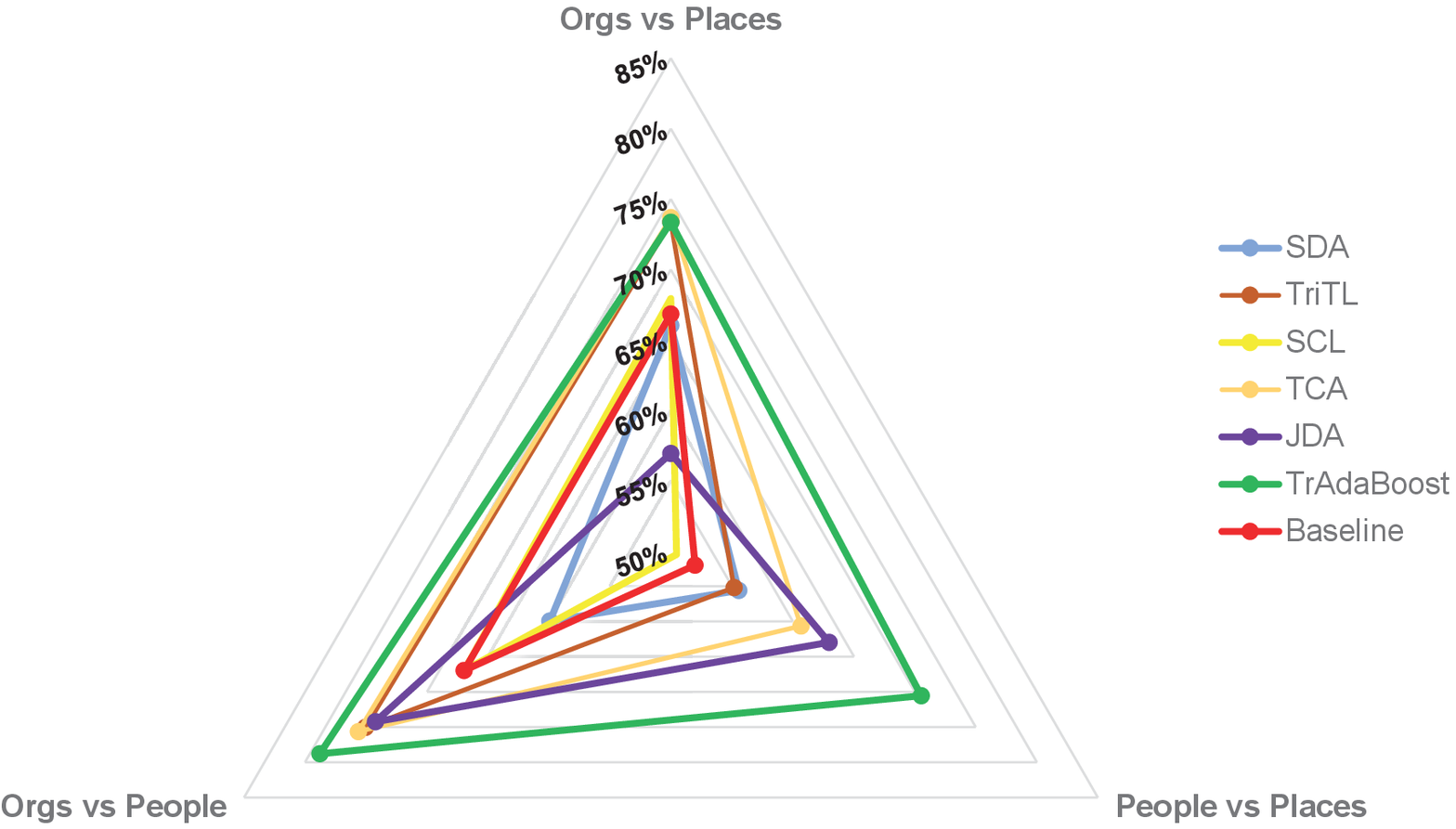}
\caption{Comparison results on Reuters-21578.}
\label{fig:reuters_results}
\end{figure*}

\begin{figure}
\centering
\includegraphics[width=0.9\linewidth]{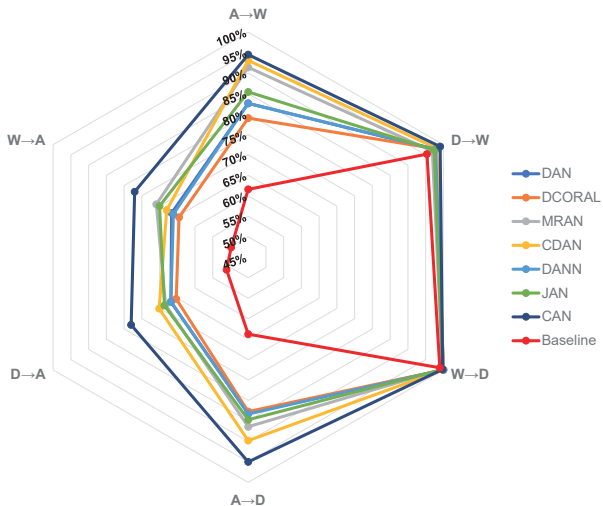}
\caption{Comparison results on Office-31.}
\label{fig:office31_results}
\end{figure}

\begin{table*}[!t]
\centering
\caption{Accuracy performance on the Amazon Reviews of four domains: Kitchen (K), Electronics (E), DVDs (D) and Books (B).}
\setlength{\tabcolsep}{2mm}{
	\begin{tabular}{cccccccccccccc}
		\toprule
		Model & K$\rightarrow$D  & K$\rightarrow$B  & K$\rightarrow$E  & D$\rightarrow$K  & D$\rightarrow$B  & D$\rightarrow$E  & B$\rightarrow$K  & B$\rightarrow$D  & B$\rightarrow$E  & E$\rightarrow$K  & E$\rightarrow$D  & E$\rightarrow$B & Average\\
		\midrule
		HIDC & 0.8800 & 0.8750 & 0.8800 & 0.7925 & 0.8100 & 0.8025 & 0.7925 &  0.8175 & 0.8075 & 0.8075 & 0.8700 & 0.8700 & 0.8338 \\
		\midrule
		TriTL & 0.7150 & 0.7250 & 0.6775 & 0.5725 & 0.5250 & 0.5775 & 0.6150 & 0.6125 & 0.6000 & 0.6250 & 0.6100 & 0.6150 & 0.6225 \\
		\midrule
		CD-PLSA & 0.7475 & 0.7225 & 0.7200 & 0.6075 & 0.6175 & 0.6075 & 0.5750 & 0.6100 & 0.6425 & 0.7225 & 0.7450 & 0.7000 & 0.6681 \\
		\midrule
		MTrick & 0.8200 & 0.8350 & 0.8125 & 0.7725 & 0.7475 & 0.7275 & 0.7550 & 0.7450 & 0.7800 & 0.7900 & 0.7975 & 0.8100 & 0.7827 \\
		\midrule
		SFA & 0.8525 & 0.8575 & 0.8675 & 0.7825 & 0.8050 & 0.7750 & 0.7925 & 0.7850 & 0.7775 & 0.8400 & 0.8525 & 0.8400 & 0.8190 \\
		\midrule
		mSLDA & 0.7975 & 0.7825 & 0.7925 & 0.6350 & 0.6450 & 0.6325 & 0.6525 & 0.6675 & 0.6625 & 0.7225 & 0.7150 & 0.7125 & 0.7015 \\
		\midrule
		SDA & 0.8425 & 0.7925 & 0.8025 & 0.7450 & 0.7600 & 0.7650 & 0.7625 & 0.7475 & 0.7425 & 0.8175 & 0.8050 & 0.8100 & 0.7827 \\
		\midrule
		GFK & 0.6200 & 0.6275 & 0.6325 & 0.6200 & 0.6100 & 0.6225 & 0.5800 & 0.5650 & 0.5725 & 0.6575 & 0.6500 & 0.6325 & 0.6158\\
		\midrule
		SCL & 0.8575 & 0.8625 & 0.8725 & 0.7800 & 0.7850 & 0.7825 & 0.7925 & 0.7925 & 0.7825 & 0.8425 & 0.8525 & 0.8450 & 0.8206 \\
		\midrule
		TCA & 0.7550 & 0.7550 & 0.7550 & 0.6475 & 0.6475 & 0.6500 & 0.5800 & 0.5825 & 0.5850 & 0.7175 & 0.7150 & 0.7125 & 0.6752 \\
		\midrule
		Baseline & 0.7270 & 0.7090 & 0.8270 & 0.7400 & 0.7280 & 0.7300 & 0.7450 &	0.7720 & 0.7080 & 0.8400 & 0.7060 & 0.7070 & 0.7449 \\		
		\bottomrule
	\end{tabular}%
}
\label{tab:result_amazon}%
\end{table*}%

\begin{table}[!t]
\centering
\caption{Accuracy performance on the Reuters-21578 of three domains: Orgs, People, and Places.}
\setlength{\tabcolsep}{0.6mm}{
	\begin{tabular}{ccccc}
		\toprule
		Model & Orgs vs Places & People vs Places & Orgs vs People & Average\\
		\midrule
		HIDC & 0.7698 & 0.6945 & 0.8375 & 0.7673 \\
		\midrule
		TriTL & 0.7338 & 0.5517 & 0.7505 & 0.6787\\
		\midrule
		CD-PLSA & 0.5624 & 0.5749 & 0.7826 & 0.6400\\
		\midrule
		MTrick & 0.7494 & 0.6457 & 0.7930 & 0.7294\\
		\midrule
		CoCC & 0.6704 & 0.8264 & 0.7644 & 0.7537\\
		\midrule
		SFA & 0.7468 & 0.6768 & 0.7906 & 0.7381\\
		\midrule
		mSLDA & 0.5645 & 0.6064 & 0.5289 & 0.5666\\
		\midrule
		SDA & 0.6603 & 0.5556 & 0.5992 & 0.6050\\
		\midrule
		GFK & 0.6220 & 0.5417 & 0.6446 & 0.6028\\
		\midrule
		SCL & 0.6794 & 0.5046 & 0.6694 & 0.6178\\
		\midrule
		TCA & 0.7368 & 0.6065 & 0.7562 & 0.6998\\
		\midrule
		JDA & 0.5694 & 0.6296 & 0.7424 & 0.6471\\
		\midrule
		TrAdaBoost & 0.7336 & 0.7052 & 0.7879 & 0.7422\\
		\midrule
		Baseline & 0.6683 & 0.5198 & 0.6696 & 0.6192\\
		\bottomrule
	\end{tabular}
}
\label{tab:reuters_results}
\end{table}

\begin{table}[!t]
\centering
\caption{Accuracy performance on Office-31 of three domains: Amazon (A), Webcam (W), and DSLR (D).}
\setlength{\tabcolsep}{0.8mm}{
	\begin{tabular}{cccccccc}
		\toprule
		Model & A $\rightarrow$ W & D$\rightarrow$W & W$\rightarrow$D & A$\rightarrow$D & D $\rightarrow$ A & W$\rightarrow$A & Average\\
		\midrule
		DAN & 0.826 & 0.977 & 1.00 & 0.831 & 0.668 & 0.666 & 0.828\\
		\midrule
		DCORAL & 0.790 & 0.980 & 1.00 & 0.827 & 0.653 & 0.645 & 0.816\\
		\midrule
		MRAN & 0.914 & 0.969 & 0.998 & 0.864 & 0.683 & 0.709 & 0.856\\
		\midrule
		CDAN & 0.931 & 0.982 & 1.00 & 0.898 & 0.701 & 0.680 & 0.865 \\
		\midrule
		DANN & 0.826 & 0.978 & 1.00 & 0.833 & 0.668 & 0.661 & 0.828\\
		\midrule
		JAN & 0.854 & 0.974	& 0.998 & 0.847 & 0.686 & 0.700 & 0.843\\
		\midrule
		CAN & 0.945 & 0.991 & 0.998 & 0.950 & 0.780 & 0.770 & 0.906\\
		\midrule
		Baseline & 0.616 & 0.954 & 0.990 & 0.638 & 0.511 & 0.498 & 0.701\\
		\bottomrule
	\end{tabular}
}
\label{tab:office31_results}
\end{table}
In this subsection, we compare over twenty algorithms on three datasets in total. The parameters of all algorithms are set to the default values or the recommended values mentioned in the original papers. The experimental results are presented in Tables \ref{tab:result_amazon}, \ref{tab:reuters_results}, and \ref{tab:office31_results} corresponding to Amazon Reviews, Reuters-21578, and Office-31, respectively. In order to allow readers to understand the experimental results more intuitively, three radar maps, i.e., Figs. \ref{fig:Amazon_results}, \ref{fig:reuters_results}, and \ref{fig:office31_results}, are provided, which visualize the experimental results. In the radar maps, each direction represents a task. The general performance of an algorithm is demonstrated by a polygon whose vertices representing the accuracy of the algorithm for dealing with different tasks.

Table \ref{tab:result_amazon} shows the experimental results on Amazon Reviews. The baseline is a linear classifier trained only on the source domain (here we directly use the results from the paper \cite{BDP2007ACL}). Fig. \ref{fig:Amazon_results} visualizes the results. As shown in Fig. \ref{fig:Amazon_results}, most algorithms are relatively well-performed when the source domain is electronics or kitchen, which indicates that these two domains may contains more transferable information than the other two domains. In addition, it can be observed that HIDC, SCL, SFA, MTrick and SDA perform well and relatively stable in all the twelve tasks. Meanwhile, other algorithms, especially mSLDA, CD-PLSA, and TriTL, are relatively unstable; the performance of them fluctuates in a range about twenty percent. TriTL has a relatively high accuracy on the tasks where the source domain is kitchen, but has a relatively low accuracy on other tasks. The algorithms TCA, mSLDA, and CD-PLSA have similar performance on all the tasks with an accuracy about seventy percent on average. Among the well-performed algorithms, HIDC and MTrick are based on feature reduction (feature clustering), while the others are based on feature encoding (SDA), feature alignment (SFA), and feature selection (SCL). Those strategies are currently the mainstreams of feature-based transfer learning.

Table \ref{tab:reuters_results} presents the comparison results on Reuter-21578 (here we directly use the results of the baseline and CoCC from papers \cite{PTK2009IJCAI} and \cite{DXY2007KDD}). The baseline is a regularized least square regression model trained only on the labeled target domain instances \cite{PTK2009IJCAI}. Fig. \ref{fig:reuters_results}, which has the same structure of Fig. \ref{fig:Amazon_results}, visualizes the performance. For clarity, thirteen algorithms are divided into two parts that correspond to the two subfigures in Fig. \ref{fig:reuters_results}. It can be observed that most algorithms are relatively well-performed for Orgs vs Places and Orgs vs People, but poor for People vs Places. This phenomenon indicates that the discrepancy between People and Places may be relatively large. TrAdaBoost has a relatively good performance in this experiment because it uses the labels of the instances in the target domain to reduce the impact of the distribution difference. Besides, the algorithms HIDC, SFA, and MTrick have relatively consistent performance in the three tasks. These algorithms are also well-performed in the previous experiment on Amazon Reviews. In addition, the top two well-performed algorithms in terms of People vs Places are CoCC and TrAdaBoost.

In the third experiment, seven deep-learning-based transfer learning models (i.e., DAN, DCORAL, MRAN, CDAN, DANN, JAN, and CAN) and the baseline (i.e., the Alexnet \cite{KSH2012NIPS,YCB2014NIPS} pre-trained on ImageNet \cite{DDS2009CVPR} and then directly trained on the target domain) are performed on the dataset Office-31 (here we directly use the results of CDAN, JAN, CAN, and the baseline from the original papers \cite{LCW2018NIPS,LZW2017ICML,KJY2019CVPR,LCW2015ICML}). The ResNet-50 \cite{HZR2016CVPR} is used as the backbone network for all these three models. The experimental results are provided in Table \ref{tab:office31_results} and the average performance is visualized in Fig. \ref{fig:office31_results}. As shown in Fig. \ref{fig:office31_results}, all of these seven algorithms have excellent performance, especially on the tasks D $\rightarrow$ W and W $\rightarrow$ D, whose accuracy is very close to one hundred percent. This phenomenon reflects the superiority of the deep-learning based approaches, and is consistent with the fact that the difference between Webcam and DSLR is smaller than that between Webcam/DSLR and Amazon. Clearly, CAN outperforms the other six algorithms. In all the six tasks, the performance of DANN is similar to that of DAN, and is better than that of DCORAL, which indicates the effectiveness and the practicability of incorporating adversarial learning.

It is worth mentioning that, in the above experiments, the performance of some algorithms is not ideal. One reason is that we use the default parameter settings provided in the algorithms' original papers, which may not be suitable for the dataset we selected. For example, GFK was originally designed for object recognition, and we directly adopt it into text classification in the first experiment, which turns out to produce an unsatisfactory result (having about sixty-two percent accuracy on average). The above experimental results are just for reference. These results demonstrate that some algorithms may not be suitable for the datasets of certain domains. Therefore, it is important to choose the appropriate algorithms as the baselines in the process of research. Besides, in practical applications, it is also necessary to find a suitable algorithm.

\section{Conclusion and Future Direction} \label{SEC.CC}
In this survey paper, we have summarized the mechanisms and the strategies of transfer learning from the perspectives of data and model. The survey gives the clear definitions about transfer learning and manages to use a unified symbol system to describe a large number of representative transfer learning approaches and related works. We have basically introduced the objectives and strategies in transfer learning based on data-based interpretation and model-based interpretation. Data-based interpretation introduces the objectives, the strategies, and some transfer learning approaches from the data perspective. Similarly, model-based interpretation introduces the mechanisms and the strategies of transfer learning but from the model level. The applications of transfer learning have also been introduced. At last, experiments have been conducted to evaluate the performance of representative transfer learning models on two mainstream area, i.e., object recognition and text categorization. The comparisons of the models have also been given, which reflects that the selection of the transfer learning model is an important research topic as well as a complex issue in practical applications.

Several directions are available for future research in the transfer learning area. First, transfer learning techniques can be further explored and applied to a wider range of applications. And new approaches are needed to solve the knowledge transfer problems in more complex scenarios. For example, in real-world scenarios, sometimes the user-relevant source-domain data comes from another company. In this case, how to transfer the knowledge contained in the source domain while protecting user privacy is an important issue. Second, how to measure the transferability across domains and avoid negative transfer is also an important issue. Although there have been some studies on negative transfer, negative transfer still needs further systematic analyses \cite{WDP2019CVPR}. Third, the interpretability of transfer learning also needs to be investigated further \cite{L2018AQ}. Finally, theoretical studies can be further conducted to provide theoretical support for the effectiveness and applicability of transfer learning. As a popular and promising area in machine learning, transfer learning shows some advantages over traditional machine learning such as less data dependency and less label dependency. We hope our work can help readers have a better understanding of the research status and the research ideas.


%



\ifCLASSOPTIONcompsoc
\section*{Acknowledgments}
\else
\section*{Acknowledgment}
\fi

The research work is supported by the National Key Research and Development Program of China under Grant No. 2018YFB1004300, the National Natural Science Foundation of China under Grant No. U1836206, U1811461, 61773361, 61836013, and the Project of Youth Innovation Promotion Association CAS under Grant No. 2017146.

\ifCLASSOPTIONcaptionsoff
\newpage
\fi



%

\end{document}